%% file: main.tex
\documentclass{article}

\PassOptionsToPackage{numbers}{natbib}

\usepackage[preprint]{neurips_2024}

\usepackage[utf8]{inputenc} 
\usepackage[T1]{fontenc}    
\usepackage{hyperref}       
\usepackage{url}            
\usepackage{booktabs}       
\usepackage{amsfonts}       
\usepackage{nicefrac}       
\usepackage{microtype}      
\usepackage{xcolor}         

\input{math_commands.tex}

\usepackage{graphicx}
\usepackage{bm}
\usepackage{enumitem}
\usepackage{amsmath}
\usepackage{amssymb}
\usepackage{cleveref}
\usepackage{float}
\usepackage{makecell}

\definecolor{indigo}{RGB}{63, 81, 181}
\definecolor{red}{RGB}{210, 40, 95} 
\definecolor{pink}{RGB}{236, 64, 122}
\definecolor{green}{RGB}{46, 182, 125}
\definecolor{blue}{RGB}{66, 133, 244}
\definecolor{yellow}{RGB}{236, 178, 46}
\definecolor{anthracite}{RGB}{13, 13, 21}

\newcommand{\chris}[1]{{\color{red}[Chris]: #1}}

\newcommand{\FA}{\textit{Feature accentuation}}

\newcommand{\pred}{\bm{f}}
\newcommand{\predl}{\bm{f}_\ell}
\newcommand{\X}{\mathcal{X}}
\newcommand{\Y}{\mathcal{Y}}
\newcommand{\A}{\mathcal{A}}
\newcommand{\F}{\mathcal{F}}
\newcommand{\T}{\mathcal{T}}
\renewcommand{\Finv}{\mathcal{F}^{-1}}
\newcommand{\tr}{\mathsf{T}}

\renewcommand{\vx}{\bm{x}}
\renewcommand{\vy}{\bm{y}}
\renewcommand{\va}{\bm{a}}
\renewcommand{\vv}{\bm{v}}
\renewcommand{\vz}{\bm{z}}
\newcommand{\btau}{\bm{\tau}}
\newcommand{\explainer}{\bm{\varphi}}

\newtheorem{theorem}{Theorem}[section]

\newtheorem{definition}[theorem]{Definition}

\usepackage{wrapfig}

\title{Feature Accentuation: Revealing 'what' \\
features respond to in natural images}

\begin{document}
\vspace{-20mm}
\maketitle

\vspace{-20mm}
\begin{center}
\textbf{Chris Hamblin}\textsuperscript{1} \footnote{email for correspondence: chrishamblin [at] fas [dot] harvard [dot] edu } \ \ \ 
\textbf{Thomas Fel}\textsuperscript{2} \ \ \ 
\textbf{Srijani Saha}\textsuperscript{1} \ \ \ 
\textbf{Talia Konkle}\textsuperscript{1} \ \ \ 
\textbf{George Alvarez}\textsuperscript{1}

$^1$Harvard University \ \ \ \ \ \ \ \ \   $^2$Brown University
\end{center}




\begin{figure}[ht]
\begin{center}
\vspace{-3mm}
\includegraphics[width=0.9\textwidth]{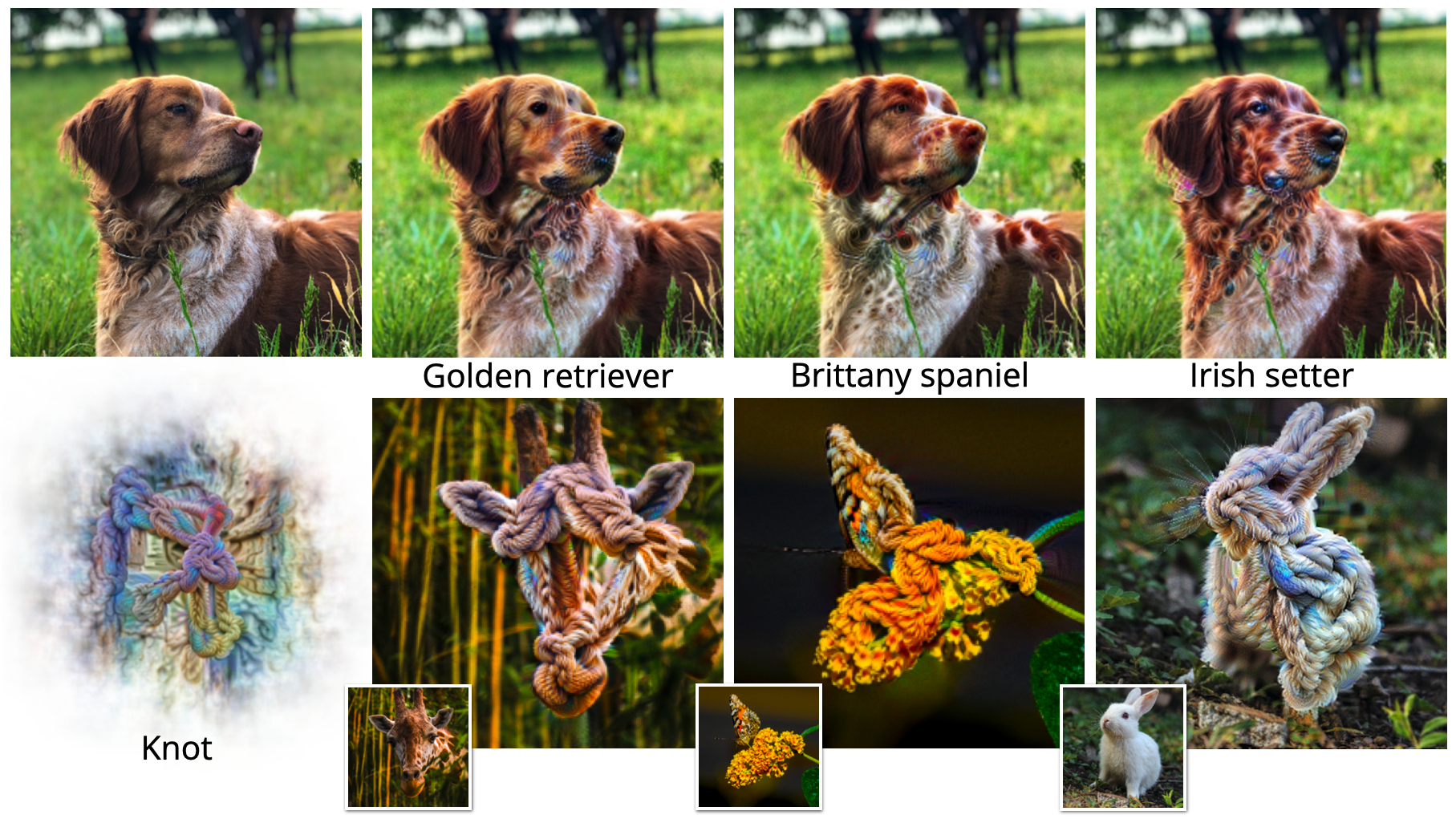}
\end{center}
\vspace{-4mm}
\caption{
%
%
We introduce \FA~, an image-seeded inputs variant of feature visualization, enabling local explanations.
(\textbf{Top}) Our method  can generate perturbations that accentuate a specific class or other logits. (\textbf{Bottom}) More generally, we propose to accentuate neurons or direction to understand model representations, (e.g. here, a knot neuron).
%
%
}\label{fig:big_picture}
\end{figure}


\begin{abstract}

Efforts to decode neural network vision models necessitate a comprehensive grasp of both the spatial and semantic facets governing feature responses within images. Most research has primarily centered around attribution methods, which provide explanations in the form of heatmaps, showing \textit{where} the model directs its attention for a given feature. However, grasping \textit{where} alone falls short, as numerous studies have highlighted the limitations of those methods and the necessity to understand \textit{what} the model has recognized at the focal point of its attention. In parallel, \textit{Feature visualization} offers another avenue for interpreting neural network features. This approach synthesizes an optimal image through gradient ascent, providing clearer insights into \textit{what} features respond to. However, feature visualizations only provide one global explanation per feature; they do not explain why features activate for particular images. In this work, we introduce a new method to the interpretability tool-kit, \textit{feature accentuation}, which is capable of conveying both \textit{where} and \textit{what} in arbitrary input images induces a feature's response. At its core, feature accentuation is image-seeded (rather than noise-seeded) feature visualization. We find a particular combination of parameterization, augmentation, and regularization yields naturalistic visualizations that resemble the seed image and target feature simultaneously. Furthermore, we validate these accentuations are processed along a natural circuit by the model. We make our precise implementation of \FA~  available to the community as the \href{https://anonymous.4open.science/r/faccent_public-0F00/}{\textit{Faccent} library}, an extension of \textit{Lucent}~\cite{lucent}.

\end{abstract}

\section{Introduction}\label{sec:intro}

Deciphering the decisions made by modern neural networks presents a significant ongoing challenge. As the realm of machine learning applications continues to expand, there is an increasing demand for robust and dependable methods to explain model decisions~\citep{doshivelez2017rigorous, jacovi2021formalizing}. 
Recent European regulations, such as the General Data Protection Regulation (GDPR)~\citep{kaminski2021right}, and the European AI Act~\citep{kop2021eu}, underscore the importance of assessing explainable decisions, especially those derived from algorithms.

In the domain of vision models, a variety of explainability methods have already been proposed in the literature~\citep{simonyan2013deep, selvaraju2017grad, smilkov2017smoothgrad, sundararajan2017axiomatic, shrikumar2017learning, zeiler2013visualizing, zeiler2014visualizing, guided-backprop, nguyen2015deep, olah2017feature, fel2021sobol, novello2022making}. One prominent category of these methods, known as \textit{attribution}, generates heatmaps that highlight significant image regions in influencing a model's decision. The heatmap can be at the level of individual pixels based on their gradients \citep{simonyan2013deep,bach2015pixel,baehrens2010explain,smilkov2017smoothgrad,sundararajan2017axiomatic,zeiler2014visualizing,springenberg2014striving,srinivas2019full,yang2020learning,montavon2017explaining}, or a coarse-grained map based on intermediate network activations/gradients \citep{zhou2016learning,Selvaraju_2019,chattopadhay2018grad,bae2020rethinking,Zhou_2018_ECCV,wang2020score,desai_2020_WACV,fuaxiom,kim2021keep} or input perturbations \citep{fel2021sobol,novello2022making,Fong_2017,zintgraf2017visualizing,petsiuk2018rise,fel2023don,ribeiro2016i,lundberg2017unified}, or even a combination of fine-grained and coarse-grained attribution \citep{Selvaraju_2019,rebuffi2020there}. 
\begin{wrapfigure}{r}{0.5\textwidth}
\begin{center}
  \includegraphics[width=.48\textwidth]{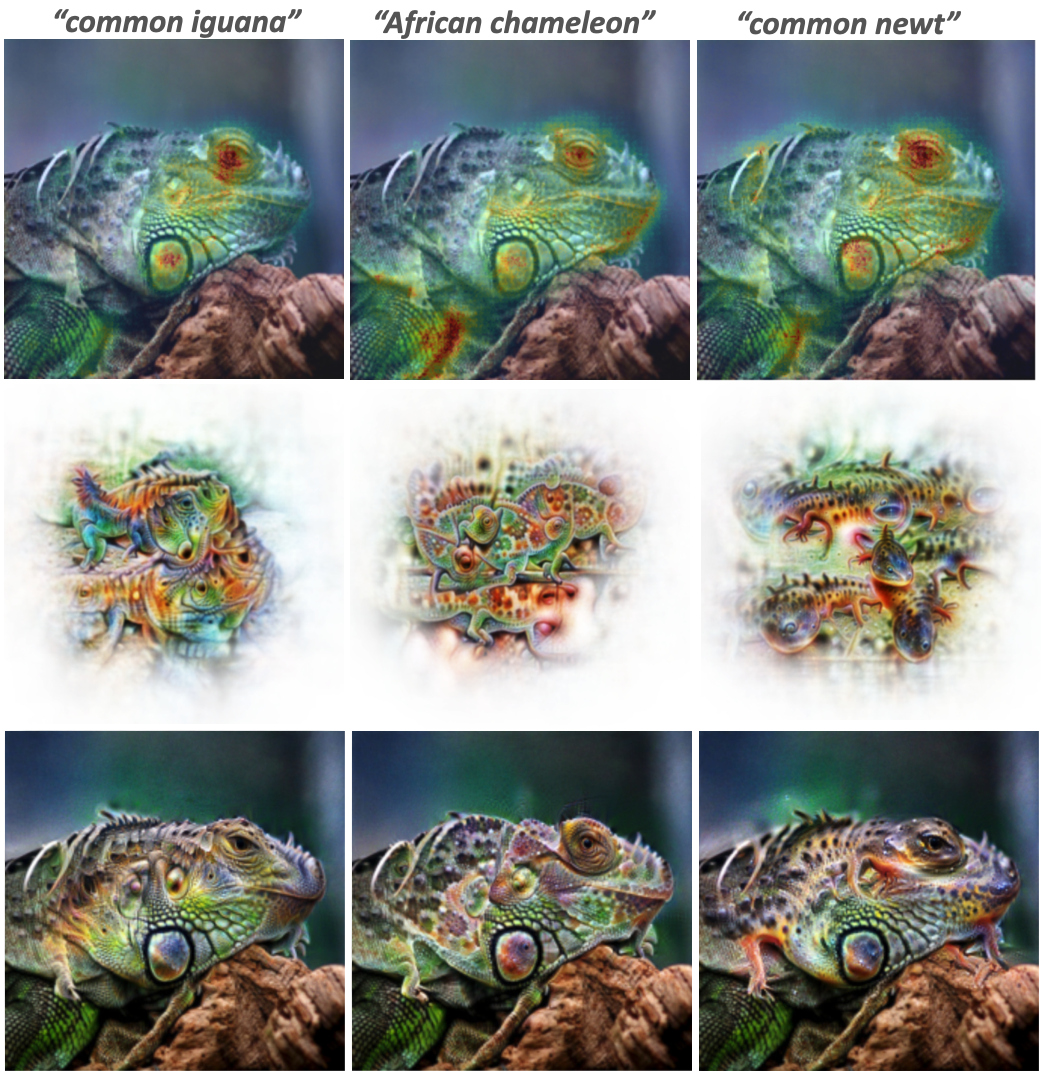}
\end{center}
\caption{An iguana excites several class logits in InceptionV1, but what about the image excites each logit? Attribution maps highlight important regions of the image, but not \textit{what} each logit sees in the region. Feature visualizations yield an exemplar for each logit, but these are hard to relate to the iguana image. Feature accentuation (ours) constitutes a powerful intermediary, transforming the iguana into a local exemplar for each class.}\label{fig:cam_versus_accent}
\end{wrapfigure}

These methods have a wide range of applications, including improving decision-making, debugging, and instilling confidence in model outputs. However, they have notable limitations and drawbacks, as highlighted in several articles~\citep{kindermans2019reliability,adebayo2018sanity,slack2020fooling,ghalebikesabi2021locality,fel2023craft,kim2021hive,hase2020evaluating,nguyen2021effectiveness,shen2020useful,sixt2022users}. These issues range from the problem of confirmation bias -- just because an explanation makes sense to a human doesn't necessarily mean it reflects the inner workings of a model -- to the more significant limitation of these methods -- they reveal \textit{where} the model is looking but not \textit{what} it has observed.
As a result, the research community is actively exploring new techniques for conveying the \textit{what} aspect of model behavior.

In response to these challenges, feature visualizations emerge as a compelling solution. They create images that strongly activate specific neurons or neuron groups \citep{szegedy2013intriguing,nguyen2015deep,olah2017feature,ghiasi2021plug,ghiasi2022vision}.
The simplest feature visualization method involves a gradient ascent process to find an image that maximizes neuron activation. However, this process can produce noisy images, often considered adversarial, without some form of control. To address this issue, various regularization techniques~\citep{olah2017feature,mahendran2015understanding,nguyen2015deep,tyka2016class,AudunGoogleNet} and data augmentations ~\citep{olah2017feature,tsipras2018robustness,santurkar2019image,engstrom2019adversarial,nguyen2016multifaceted,mordvintsev2015inceptionism} have been proposed, along with the use of generative models~\citep{wei2015understanding,nguyen2016synthesizing,nguyen2017plug}.
%

Despite the promise of these methods, recent research has pointed out their limitations and potential pitfalls~\citep{zimmermann2021well,geirhos2023dont}. It is argued feature visualizations may be misleading, as they can trigger intermediate network features that differ significantly from natural images; i.e. they produce neuron activation by way of a different circuit. Additionally, a given neuron may return high activation for a broad range of images, many of which cannot be easily related to the exemplars generated with feature visualization. In many practical settings, it is necessary to explain why a neuron responds to some \textit{particular natural image}, such as an image that is misclassified. Thus in practice attribution methods see far more use than feature visualizations, given they provide image conditional explanations.\par
A final family of techniques that could serve to explain these difficult edge cases are visual counterfactual explanations (VCEs)~\citep{goyal2019counterfactual,poyiadzi2020face,verma2020counterfactual}. VCEs attempt to specify an image \(\bm{x}'\) close to \(\bm{x}\) that explains '\textit{how would} \(\bm{x}\) \textit{need to change for the model to consider it an instance of class} \(\bm{y}\)\textit{?}'  Contemporary VCE methods utilize variations of image-seeded (rather than noise-seeded) feature visualization, thus could potentially be used to explain \textit{what} excites a feature in a particular natural image. However, visually compelling VCEs have only been synthesized in this way using either adversarially trained models~\citep{boreiko2022sparse,gaziv2023robustified,madry2017pgd} or by optimizing under the guidance of an auxiliary generative model~\citep{augustin2022diffusion}. Thus far in the literature, image-seeded feature visualizations of non-robust models under self-guidance yield one of two results, neither of which constitute a useful explanation. 1) When optimization is unconstrained, unnatural \textit{hallucinations} appear across the entire image, similar to those popularized by \textit{deepdream}~\citep{mordvintsev2015inceptionism}. 2) When the optimization is constrained to remain close to the seed image, the synthesized image does not change in a perceptually salient way, as with classic adversarial attacks~\citep{szegedy2013intriguing,goodfellow2014explaining,moosavi2016deepfool,akhtar2018threat}.\par

In the present work we introduce \textit{Feature Accentuation}, a novel implementation of image-seeded feature visualization that does not require robust/auxiliary models.
 Feature accentuation can provide diverse explanations across both neurons and the images that activate them. Specifically, in this article:

\begin{itemize}
    \item We formally introduce Feature Accentuation, a new post-hoc explainability method that reveals both \textit{where} and \textit{what} in natural images induces neuron activation.
    \item We show for the first time that high quality VCEs can be generated for non-robust models with image parameterization, augmentation, and regularization alone, without the use of an auxilary generative model.
    \item We show that feature accentuations drive neurons through the same circuits as natural images, which is not true of conventional feature visualizations. In fact, feature accentuations follow a class logit's prototypical circuit \textit{more closely} than natural images do!
    \item We demonstrate the diverse range of applications feature accentuation facilitates, and validate its utility with a human experiment.
    \item We release the open-source \textit{Faccent} Library, an extension of the \textit{Lucent} \citep{lucent} feature visualization library for generating naturalistic accentuations.
\end{itemize}

\vspace{-3mm}
\section{Methods}
\paragraph{Hardware \& Software.} The following experiments were conducted with the aid of 2 GeForce RTX 2080 GPUs. We implemented our experiments in Pytorch (BSD 3-Clause License) \cite{paszke2019pytorch}, and leveraged code from the Lucent (Apache License 2.0) \cite{lucent} feature visualization library.

\paragraph{Notation.} 
Throughout, we consider a general supervised learning setting, with an input space $\X \subseteq \mathbb{R}^{h \times w}$, an output space $\Y \subseteq \mathbb{R}^c$ and a classifier $\pred : \X \to \Y$ that maps inputs $\vx \in \X$ to a prediction $\vy \in \Y$.
Without loss of generality, we assume that $\pred$ admits a series of $L$ intermediate spaces $\A_\ell \subseteq \mathbb{R}^{p_\ell}, 1 < \ell < L$. 
In this setup, $\predl : \X \to \A_\ell$ maps an input to an intermeditate activation $\va = (a_1, ..., a_{p_\ell})^\tr \in \A_\ell$ of $\pred$.
We respectively denote $\F$ and $\Finv$ the 2-D Discrete Fourier Transform (DFT) on $\X$ and its inverse.
In this article, a feature denotes a vector in a feature space $\vv \in \A_\ell$ and its associated feature detector function is denoted as $\pred_{\vv}(\vx) = \predl(\vx) \cdot \vv$.
Finaly, $|| \cdot ||$ denotes the $\ell_2$ norm over the input and activation space. 
%


\paragraph{Feature Visualization.} Activation maximization attempts to identify some prototypîcal input $\vx^{\ast} = \argmax_{\vx} \pred_{\vv}(\vx)$ for any arbitrary feature detector function $\pred_{\vv}$ of interest (e.g. a channel or a logit of a convolutional neural network). 
To generate meaningful images, as opposed to noisy ones, a variety of techniques are typically employed, including some that we will also adopt in our approach. First and foremost, the choice of image parametrization plays a critical role, which is typically not pixel space but the frequency domain $\mathcal{Z} \subseteq \mathbb{C}^{h \times w}$, which give us $\vx = \Finv(\vz)$. This parameterization affords precise control over the image's frequency components, facilitating the removal of adversarial high-frequencies. Additionally, this re-parametrization has the effect of modifying the basins of attraction during optimization, yielding images that exhibit a more natural appearance. A second commonly employed strategy is the application of transformations at each gradient step to enhance image robustness. Noise injection or localized image optimization through cropping are frequently employed in this context. Formally, we denote $\T$ as the set of possible transformations, and $\btau \sim \mathcal{T}$ represents a transformation applied to $\vx$ such that $\btau(\vx) \in \X$.
Putting everything together, for a feature detector $\pred_{\vv}$, we associate the optimal feature visualization $\vz^{\ast}$ :

$$
\vz^{\ast} = \argmax_{\vz \in \mathbb{C}^{h \times w} } \mathcal{L}(\vz) 
~~ with ~~ \mathcal{L}(\vz) \triangleq (\pred_{\vv} \circ \btau \circ \Finv)(\vz)
$$

However, this feature visualization method presents several challenges: it lacks image specificity and therefore fails to provide local explanations (for a given point). Most of the time, it converges to a perceptually similar optima, providing only one visual explanation of the feature detector. Finally, as we will explore in the Section~\ref{sec:path}, \cite{geirhos2023dont} observe that the generated images follow circuits which are very different from the circuits followed by natural images, thus raising questions about the validity of the explanations. We address these issues by introducing \FA.

\subsection{Feature accentuation}

\begin{figure}[ht]
\begin{center}
  \includegraphics[width=\textwidth]{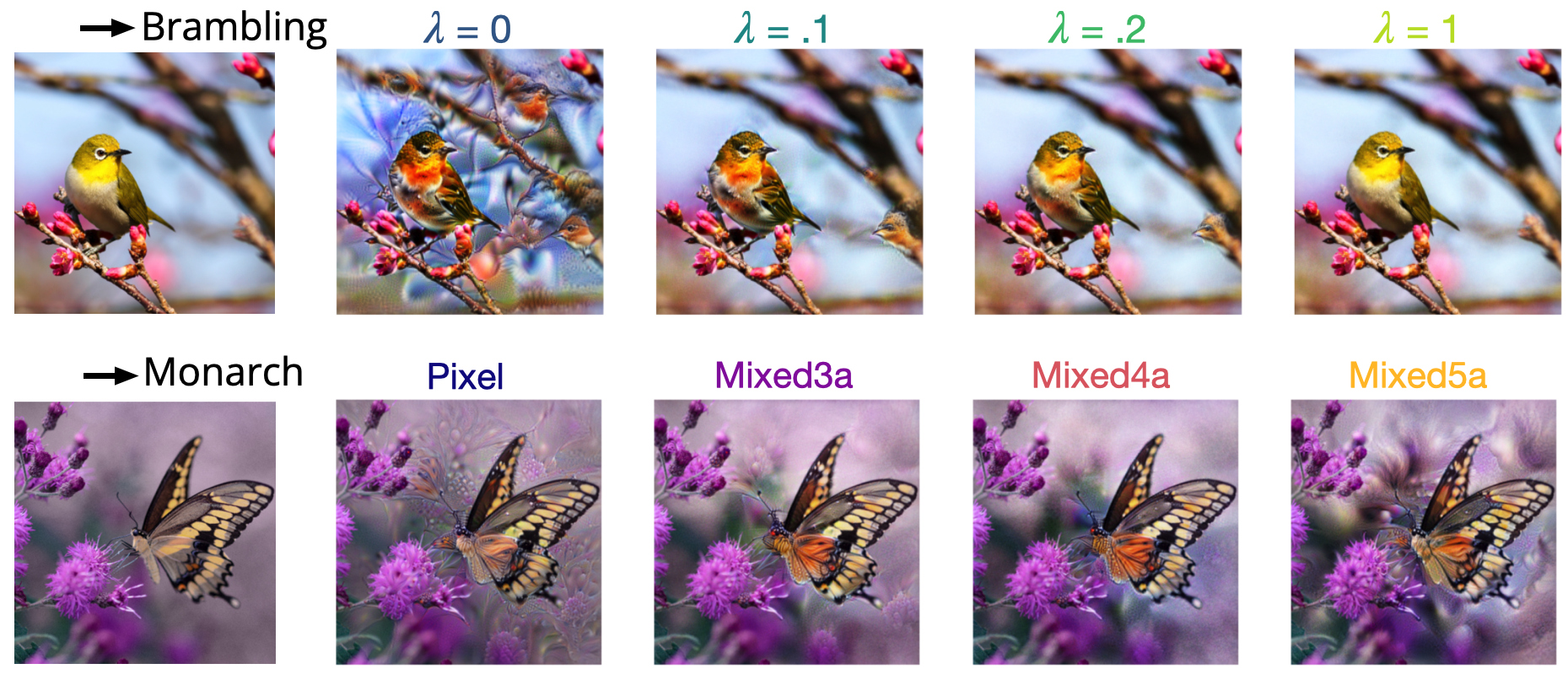}
\end{center}
\caption{
\textbf{(Top)} Influence of Regularization Strength ($\lambda$). With no regularization accentuations significantly deviate from the original image, while excessive regularization prevents meaningful alterations to the image.
\textbf{(Bottom)} Influence of Regularization Layer ($\predl$) -- from pixel space ($\X$) to a deep layer of InceptionV1 (mixed5). Regularization in pixel space does not enable meaningful image modifications, whereas regularization in excessively deep layers produces hallucinations.
Intermediate regularization accentuates existing image details that drive the logit without injecting new features across the entire image.
}
\label{fig:adherence_qual}
\end{figure}

We begin by adapting feature visualization through seeding, i.e., initializing this process at $\vz_0 = \Finv(\vx_0)$, where $\vx_0$ represents the natural image under investigation. In addition to introducing diversity into the explanations -- each image will have a really different accentuation, even for similar neuron maximization -- this approach allows us to commence from a non-OOD (Out-Of-Distribution) starting point. However, there is no guarantee that during the optimization process the image will remain natural or be perceived as such by the model. Additionally, if one desires a local explanation of \(\pred_{\vv}(\vx_0)\), it is possible the synthesized image is processed differently by the model than the seed image. To address these concerns, we will augment the loss function with a regularizer that encourages intermediate activation to be closer to the original one. Formally, we introduce \FA~loss:

\begin{definition}[\textbf{\FA}]\label{def: fa}
The feature accentuation results from optimizing the original image $\vx_0$ such that:


$$
\vz^\ast = \argmax_{\vz \in \mathbb{C}^{h \times w} } 
    \underbrace{(\pred_{\vv} \circ \btau \circ \Finv)(\vz)}_{\text{maximize feature activation}} - 
    \overbrace{\lambda || (\predl \circ \btau \circ \Finv)(\vz) - (\predl \circ \btau)(\vx_0) ||}^{\text{maintain proximity to the \textbf{transformed} original image}} 
$$
\end{definition}


In various image manipulation applications, the regularization term is often defined in terms of pixel space distance \citep{santurkar2019image,augustin2020adversarial,boreiko2022sparse}. However, for our specific application, our concern is not the pixel space distance per se, but rather ensuring that the synthesized image is processed in a manner similar to the seed image by the model. Consequently, we explore the implications of regularizing based on distances measured in the latent layers of the model. Here, we draw inspiration from previous work on "feature inversion" \citep{mahendran2015understanding,olah2017feature}, in which an image is synthesized to possess the same latent vector as a target image. The choice of the layer for feature inversion significantly impacts the perceptual similarity between the synthesized image and the target. Earlier layers tend to produce synthesized images that are more perceptually similar, as these layers share more mutual information with the pixel input. Conversely, using later layers introduces differences in the synthesized image to which the higher-order representations are invariant.
We note that a distinctive feature of our approach compared to previous works is that we apply the same augmentation $\btau$ to both the target $\vx_0$ and the accentuated image $\Finv(\vz)$ at each iteration, ensuring a robust alignment with the original image.
Figure \ref{fig:adherence_qual} illustrates the impact of $\lambda$ on the feature accentuation process. A smaller value of $\lambda$ permits substantial alterations across the entirety of the image. In contrast, a larger $\lambda$ leads to accentuation that closely resembles the seed image. The choice of the layer for measuring distance also exerts an influence on the outcome. Traditional pixel space regularization tends to produce subtle, low-amplitude patterns distributed across the entire image. Conversely, regularization in the latent space of the model both constrains and enhances the accentuation to critical regions. We believe this phenomena can be understood as the edge detectors early in the model (and in our own visual system) boosting low amplitude pixel differences (appendix \ref{sec:pixel space app}).

%
%
We find that applying regularization in the early layers of the model yields comparable and desirable outcomes, whereas regularization in later layers leads to distortions resembling those observed in prior work on feature inversion.

\subsection{The appropriate augmentation}

\begin{figure}[h]
\begin{center}
  \includegraphics[width=\textwidth]{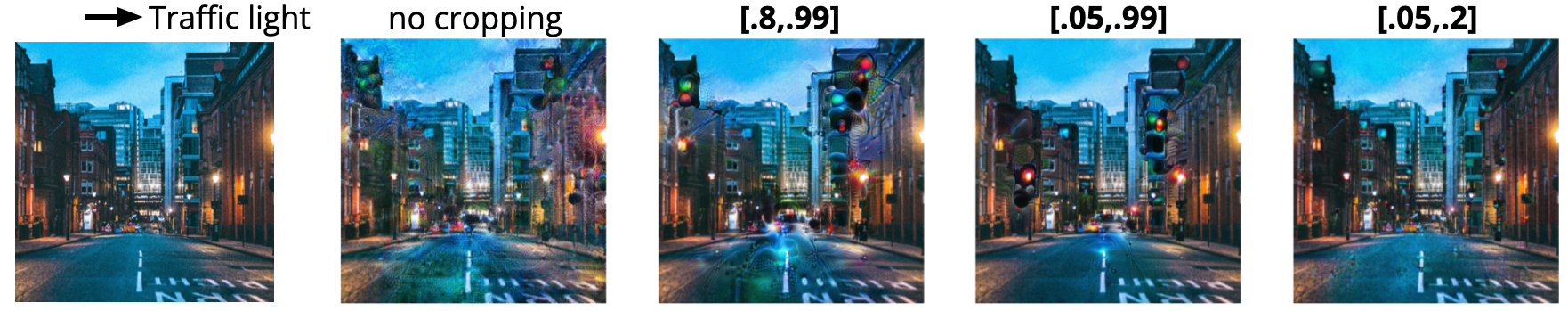}
\end{center}
\caption{The effect of random \textbf{crop augmentations}. 
Square brackets indicate the \texttt{[minimum, maximum]} permissible dimensions (in \%) for the bounding box crop. Smaller crops add definition to the image, but when only small crops are applied tiny hallucinations appear. We find applying the full range of crops each batch to produce balanced results.
}\label{fig:cropping_qual}
\end{figure}

\cite{fel2023maco} show that random cropping  with gaussian and uniform noise are a simple yet effective set of augmentations for feature visualizations. Where the \textit{MACO} crop augmentations were sampled with a center bias in the original work, we sample uniformly for feature accentuations. We apply 16 random transformations every iteration, averaging $\mathcal{L}$ over this batch. We find that incorporating smaller box crops into this augmentation scheme improves the definition of feature accentuations, as can be seen in Figure \ref{fig:cropping_qual}. 

%
%

We also observe that image parameterization has important effects on feature accentuations. We test a variety of options proposed in literature in the appendix (section \ref{sec: parameterization}), and based on this analysis recommend the Fourier parameterization \cite{olah2017feature,mordvintsev2018differentiable} for most applications.

\subsection{Adding spatial attribution}
\label{sec:masking}

\begin{figure}[h]
\begin{center}
  \includegraphics[width=\textwidth]{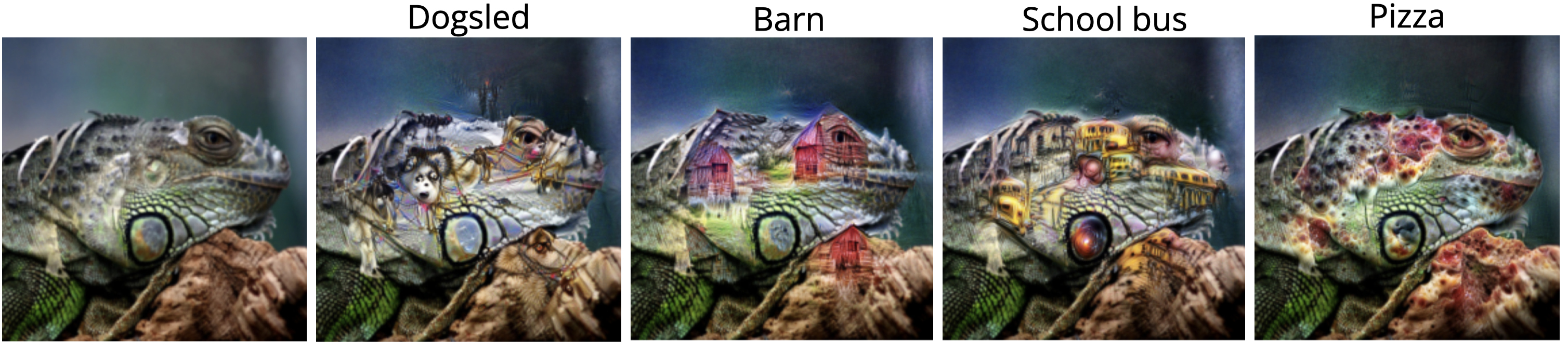}
\end{center}
\caption{\textbf{Accentuating unrelated features in images can lead to significant changes.} Accentuations of the most inhibited logits for the iguana image. While the model doesn't associate any 'schoolbus-like' features with the iguana, an observer might mistakenly think otherwise due to the result of feature accentuation. To address this, we suggest incorporating spatial attribution in Section \ref{sec:masking} as the final ingredient  for \FA.
}
\label{fig:misleading}
\end{figure}


We have demonstrated that a combination of regularization, parameterization, and augmentation techniques can produce naturalistic enhancements of image features, which exhibit similarities to both the original seed image and the target feature.
However, let us revisit the primary objective of our methodology: feature accentuations should elucidate \textit{what} a feature responds to "within the seed image." What conclusions might one draw from accentuations in this regard? 
Intuitively, \FA~ maximizes activation by \textit{exaggerating} important areas of the seed image. Thus, one might conclude the feature is present in the seed image in precisely those areas that undergo change.

Afterall, we have seen that with the correct implementation  presumably unimportant regions of the image, such as the background, experience minimal perceptual changes when accentuated. However, it is crucial to observe that an image already exemplifying the target feature will also undergo no changes when accentuated, since it already represents a local optimum. 
\begin{wrapfigure}{r}{0.4\textwidth}
  \centering 
 \includegraphics[width=0.38\textwidth]{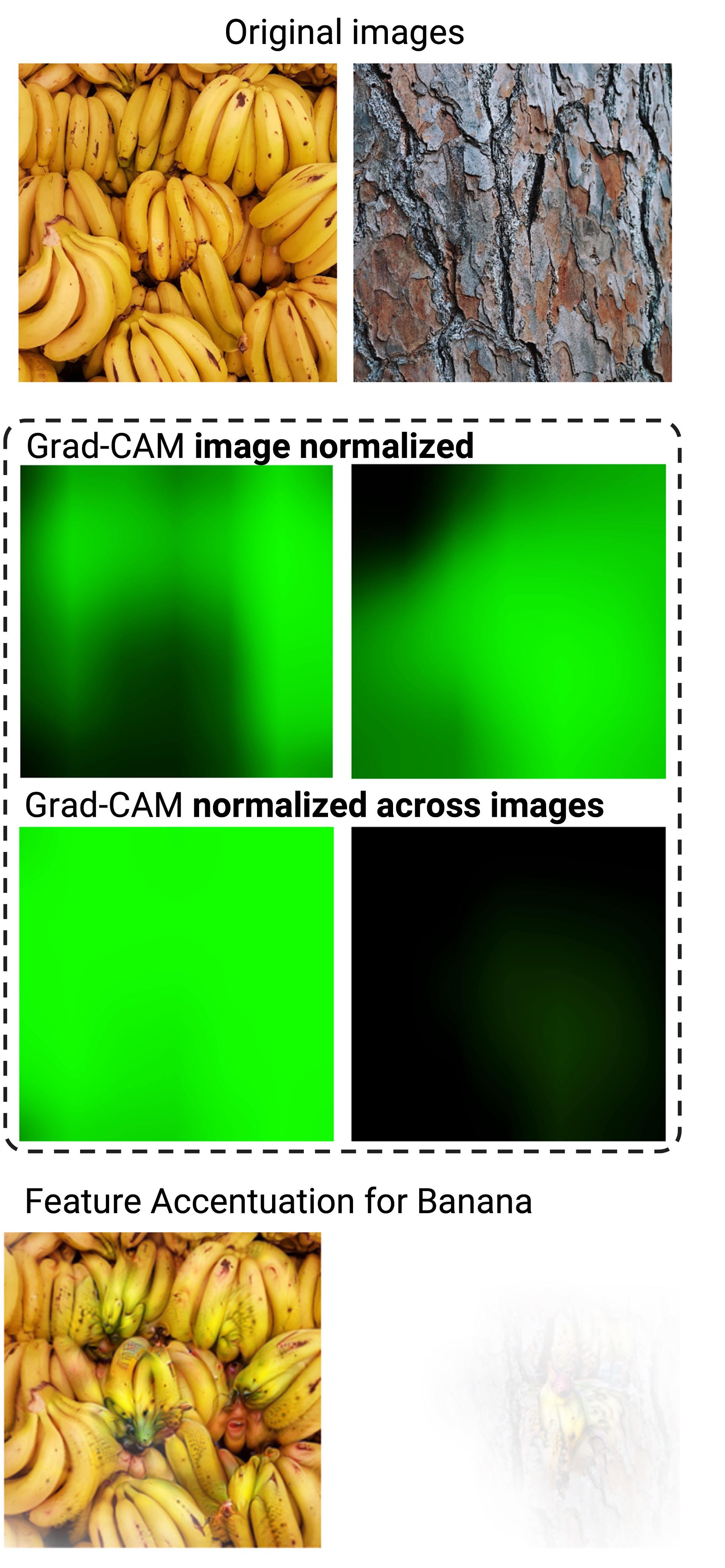}
  \caption{We introduce a normalization \textit{across} images to provide a sensible mask for \FA.
  }
  \label{fig:mask}
\end{wrapfigure}

Conversely, images unrelated to the target feature exhibit substantial changes upon accentuation. For instance, Figure \ref{fig:misleading} illustrates accentuations of the logits \textit{most inhibited} by the iguana image. Notably, the model does not associate any "schoolbus-like" attributes with the iguana, but an observer might erroneously interpret otherwise based on the corresponding feature accentuation. In light of this consideration, we propose that the final component of \FA~should encompass spatial attribution, thereby reintroducing \textit{where} into the \textit{what}.

As discussed in Section~\ref{sec:intro}, numerous attribution methods have been introduced in the literature. In this article, we maintain a general perspective, not favoring any specific algorithm. However, for an attribution map—computed by any means—to align effectively with the objectives of feature accentuation, we propose a straightforward modification.

In the typical scenario, an attribution map is represented as a function $\explainer(\pred_{\vv}, \vx) \subseteq \mathbb{R}^{h \times w}$, where the values in the attribution map $\explainer(\pred_{\vv}, \vx)$ delineate significant spatial regions in $\vx$ that influence the activation produced by the feature detector. Visualizing an attribution as a heatmap necessitates normalization. Typically, normalization occurs with respect to the values in a single attribution map, rendering the intensity entirely contingent on the variance of  $\explainer(\pred_{\vv}, \vx)$. 
%
%
As a result, an image containing prominent features, such as bananas distributed extensively, might undergo normalization that diminishes the impact of these features, leading to an explanation that does not effectively capture their significance. Conversely, an image devoid of such features may experience a normalization process that artificially amplifies any minor attributes present. Consequently, both images can end up with similar explanations, despite substantial differences in their content and the features of interest (as shown in Figure \ref{fig:mask}).

To mitigate this issue and improve the fidelity of feature representations, we propose a ``collective'' normalization approach, which leverages the attribution values derived from a comprehensive set of natural images. This approach enables a more accurate assessment of feature relevance across diverse contexts, allowing for a finer-grained distinction between images with disparate feature compositions.
Subsequently, these resulting maps can serve as opacity masks for the feature accentuation applied to the image.
%
%
As illustrated in Figure ~\ref{fig:mask}, this normalization approach enables the possibility that an image may activate a feature throughout its entirety or not at all.

\vspace{-2mm}
\section{Experiments}\label{sec:experiments}
\vspace{-2mm}

\label{sec:path}

\begin{figure}
   \centering
   \includegraphics[width=\textwidth]{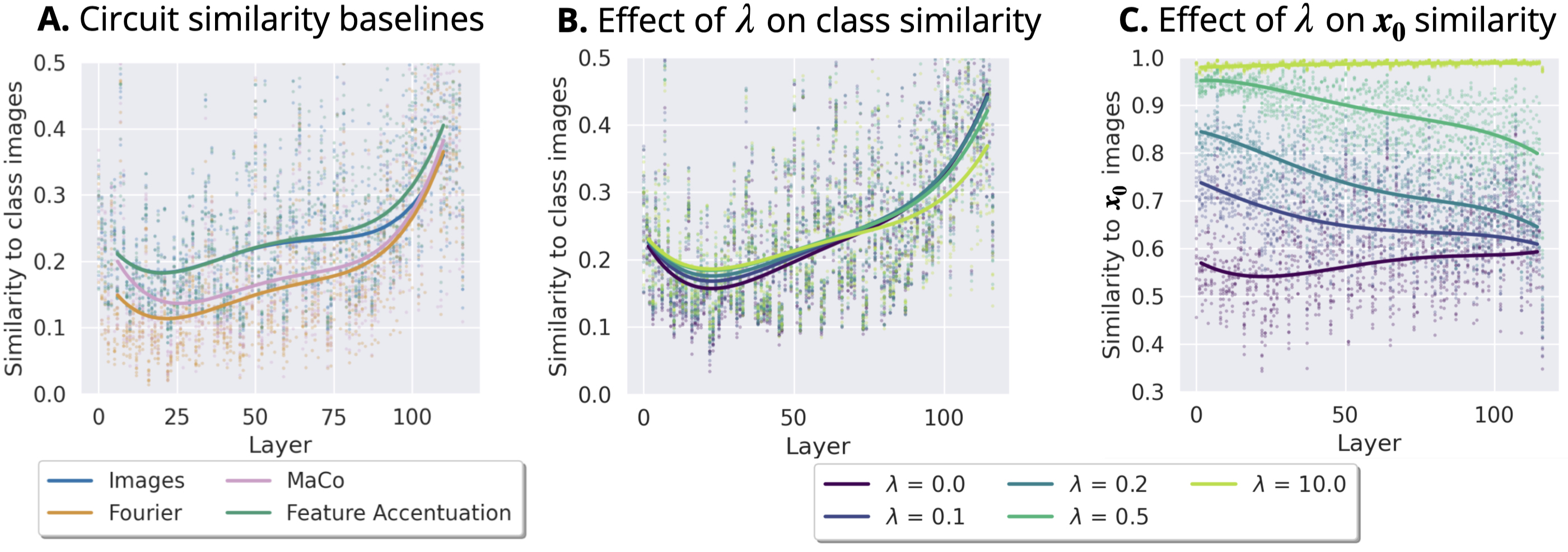}
   \caption{\textbf{Path similarity of Feature Accentuations.} \textbf{A.} We measure path similarity (correlation) of accentuations and natural images,  as compared to the similarity of other synthesized/natural images. Unlike traditional feature visualization, \FA~ for a class follow an internal path that is similar to natural images of that class,  and even closer on average to natural images than the natural images themselves, hence the term ``supernatural images''
   \textbf{B.} We observe the impact of \(\lambda\) on this measure, and  \textbf{C.} 
   how the path similarity to the seed image is influenced by $\lambda$. A high $\lambda$ value brings \FA~ closer to the original image, while a $\lambda$ value of zero results in deviation from naturalistic images towards something resembling feature visualization. This reveals a ``sweet spot'' for $\lambda$ in achieving supernatural images
   }
   \label{fig:path_experiment}
\end{figure}

\paragraph{Circuit Coherence Assessment}

 \FA~ produces perceptually meaningful transformations, but if they are to constitute good explanations of a model's responses, it's important we validate they are processed by the model in a natural way. To assess this, we adopt an approach introduced in \cite{geirhos2023dont}, which measures the degree to which the paths (intermediate activations) taken by the images align with natural images of the studied class.
Formally, to quantify this concept of ``circuit similarity'' between two images, $\vx_{i}$ and $\vx_{j}$, we measure the pearson correlation of their hidden vectors, $\rho(\va_{\ell, i}, \va_{\ell, j})$, and average across pairs of images, across all layers $\ell$ of the network. As depicted in Figure \ref{fig:path_experiment} (A), conventional feature visualization methods perform inadequately in this regard. Specifically, it has been demonstrated that for natural images of a class and his associated feature visualizations, there exists a low correlation across most layers of the network when compared to the correlation observed among natural images themselves.
In contrast, \FA~capitalizes on its image-seeded approach and regularization techniques to maintain a notably high level of naturalness in its internal pathways. It even achieves a super-natural score, signifying a superior inter-correlation with natural images than the correlation observed among natural images themselves.

\paragraph{Effect of $\lambda$.} Based on these findings, it is pertinent to investigate the behavior of the pathway in function of the introduced regularization parameter, $\lambda$. Figure \ref{fig:path_experiment} (B) illustrates the effect of varying $\lambda$ on the naturalness of the pathway. Naturally, we observe that strong regularization compels the image path to remain close to the natural images, resulting in excellent scores for early layers but deteriorating scores in later layers. Conversely, moderate regularization allows us to maintain significantly higher scores compared to unregularized optimization, suggesting the existence of an optimal $\lambda$ value.
Furthermore, in Figure \ref{fig:path_experiment} (C), we examine the correlation between the optimized image and, instead of the distribution of natural images, the seed image. As anticipated, a higher $\lambda$ corresponds to a higher correlation. It is noteworthy that, to achieve a strong score in class similarity, \FA~ does not necessarily adhere faithfully to the input image. Instead, it employs pathways that exist in other images.

\paragraph{Misclassifications}

\begin{figure}[h]
\begin{center}
  \includegraphics[width=\textwidth]{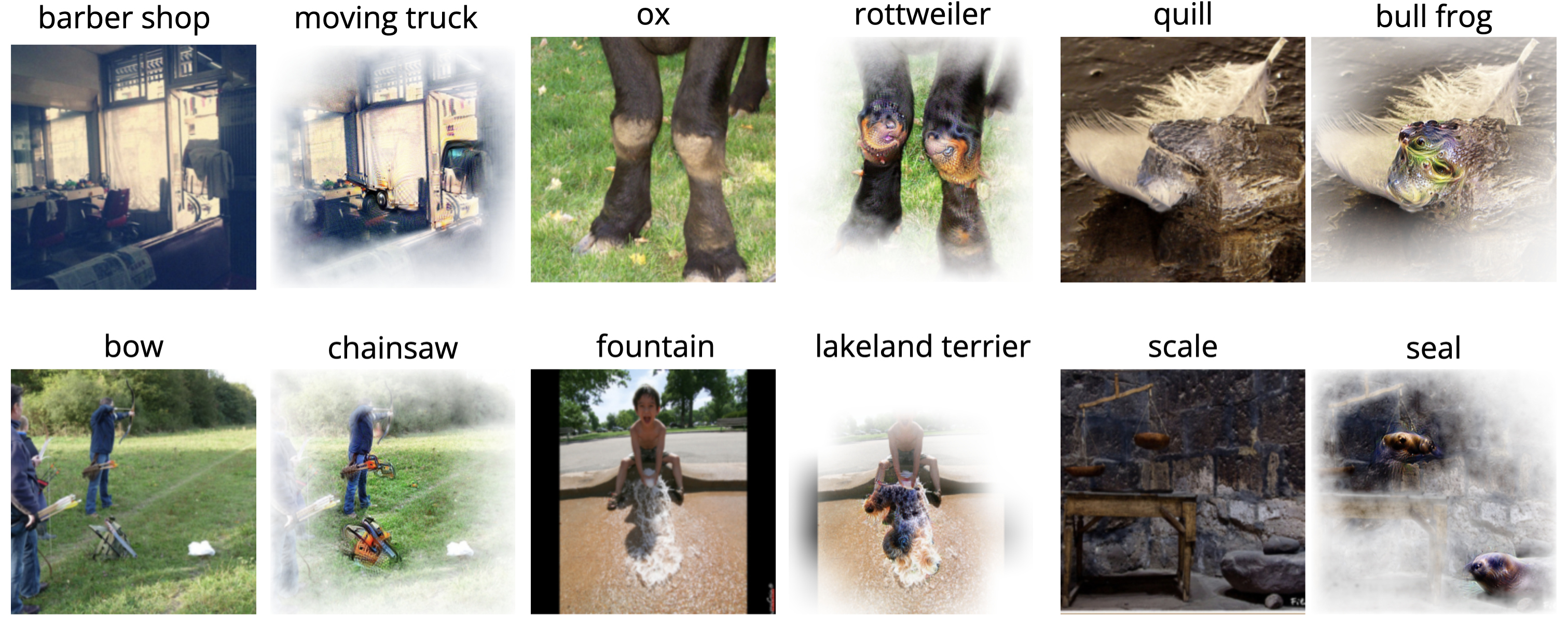}
\end{center}
\caption{\textbf{\FA~enables understanding of failure cases} by emphasizing incorrectly predicted classes, allowing the human user to share in the model's hallucination. For instance, the top left example shows a barber shop misclassified as a truck due to the shop's wrinkled curtain being mistaken for a truck's side, while the first example in the bottom row displays an archery image classified as a chainsaw because the orange-tipped arrows resemble a the handle, and the gray shafts resemble a saw blade.
}\label{fig:misclassification}
\end{figure}

Understanding why a model misclassifies inputs is one of the primary aims of explainability research. Figure \ref{fig:misclassification} demonstrates how exaggerating the predicted logit with \FA can assist in diagnosing misclassifications, helping the user see the same hallucination as the model. 



\paragraph{Explaining Latent Features}

A comprehensive understanding of a neural network model must include its latent features in addition to its logits. However, explaining activations within the latent space poses a greater challenge than explaining activations in the logit space due to the absence of direct correspondence between latent features and predefined semantic labels. Typically, latent features are probed using feature visualizations or dataset examples that elicit the highest activations. 
Feature accentuations present a novel approach for unveiling the multifaceted manifestations of a specific latent feature within natural images. Similar to conventional feature visualizations, this process can be executed with respect to individual neurons, channels, or even arbitrary feature directions, such as those associated with conceptual dimensions as in~\citep{fel2022craft}.

\begin{wraptable}{r}{0.5\textwidth}
  \centering
  \begin{tabular}{|c|c|c|}
    \hline
    \makecell{Explanation \\ Type} & \makecell{Mean Accuracy \\ \(\pm\) SE (\%)} & \makecell{\# Best \\ Trials} \\
    \hline
    None & \(50.3 \pm 5.05\) & 17 \\
    Dataset & \(55.9 \pm 5.04\) & 14 \\
    Accentuation & \(58.4 \pm 4.98\) & 22 \\
    \hline
  \end{tabular}
  \caption{Human performance across explanation types}
  \label{tab:wraptable}
\end{wraptable}
\paragraph{Human Experiment} It has been argued extensively that a good explanation is one that allows the user to extrapolate, i.e. better predict model behavior in novel scenarios \cite{doshivelez2017rigorous,shen2020useful,nguyen2022visual,kim2021hive,sixt2022users}. Here we adopt the extrapolation paradigm to the use case for which feature accentuation is designed -- local explanations for feature activations that are atypical/unintuitive. 
We compare our technique to maximally activating dataset examples, which have been shown to be more effective for extrapolation than conventional feature visualizations \cite{borowski2020exemplary,zimmermann2021well}. However, it's possible that a given dataset example is not sufficiently local, activating the target feature for a different reason than the image it is meant to explain. To test this, we compare feature accentuations and top activating dataset examples as explanations of images that yield large activations for a target feature, but are nonetheless cosine dissimilar from the those dataset examples in the latent space for the feature. We then test participants on their ability to locally extrapolate in a binary choice test, selecting between two images that are cosine similar to the original, but only one of which yields a large activation for the target feature. We find in this experiment that feature accentuation explanations yield better extrapolation on average across features (\(p=.021\) in a two sample t-test), as well as more individual features for which its explanation is best. Comprehensive details for this experiment can be found in appendix \ref{sec:human experiment}.


\begin{figure}[h]
\begin{center}
  \includegraphics[width=\textwidth]{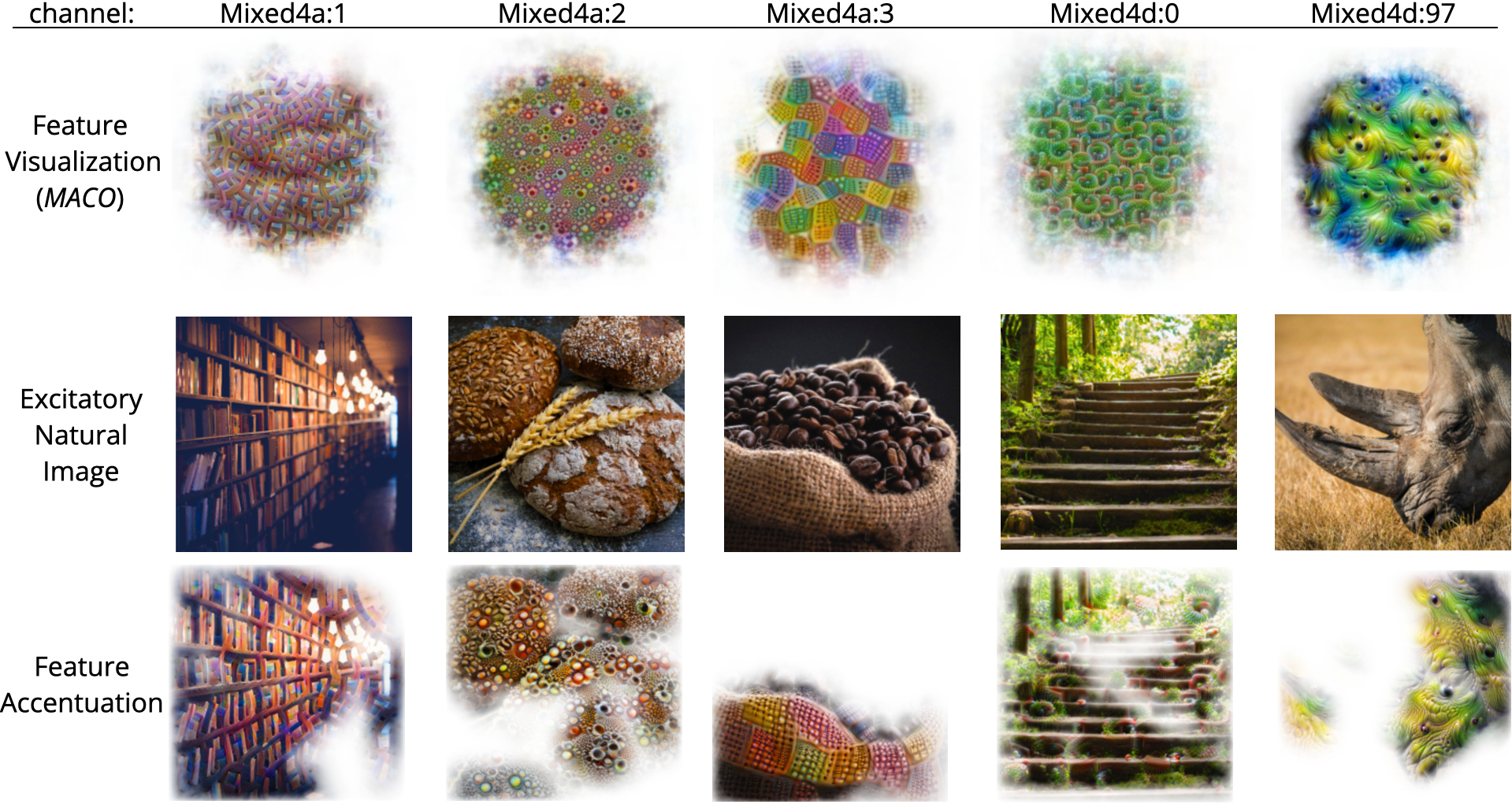}
\end{center}
\caption{\textbf{\FA~applied to latent channels} can reveal the multitude of ways features manifest within natural images. In contrast to conventional feature visualization, feature accentuation allows us to amplify the presence of targeted features in real-world images, granting insights into how they \textit{behave in the wild}.
}\label{fig:latents}
\end{figure}

\vspace{-2mm}
\section{Limitations}
\vspace{-2mm}

We have demonstrated the generation of realistic feature accentuations for neural networks using an improved feature visualization technique. However, it is essential to recognize that creating realistic images does not automatically guarantee effective explanations of neural networks. 
Furthermore, to gain informative insights into the model, especially in identifying spurious features, feature visualizations might need to produce images that deviate from the original input, potentially conflicting with the proposed regularization.
Previous research has also shed light on the limitations of feature visualizations, as observed in well-regarded studies~\citep{borowski2020exemplary,geirhos2023dont,zimmermann2021well}. A prominent critique concerns their limited interpretability for humans. Research has shown that dataset examples are more effective than feature visualizations for comprehending CNNs. This limitation could arises from the lack of realism in feature visualizations and their isolated application.
We strongly advocate for the use of supplementary tools such as attribution maps and concept-based explanations alongside Feature Accentuation, to build a comprehensive understanding of neural networks.

\vspace{-2mm}
\section{Conclusion}
\vspace{-2mm}

Understanding why a model detects a particular feature in an image is a pivotal challenge for XAI. In this article, we introduce \textit{Feature Accentuation}, a tool designed to exaggerate features in an image to aid our understanding. This results in meaningful and natural morphisms without resorting to GANs or robust models, thereby ensuring that the explanations provided by \FA~ are exclusively attributable to the scrutinized model.
Our findings demonstrate that the generated images (i) undergo processing akin to natural images, (ii) can help comprehension of model failure cases and (iii) internal features. We anticipate that the utilization of \FA~ will pave the way for enhanced XAI methodologies and better understanding of the internal representation of discriminative vision models.

\bibliography{iclr2024_conference}
\bibliographystyle{iclr2024_conference}



\newpage
\appendix
\section*{\hfil ~~~~ Appendix for Feature Accentuation \hfil}


\begin{figure}[H]
   \centering
   \includegraphics[width=0.49\textwidth]{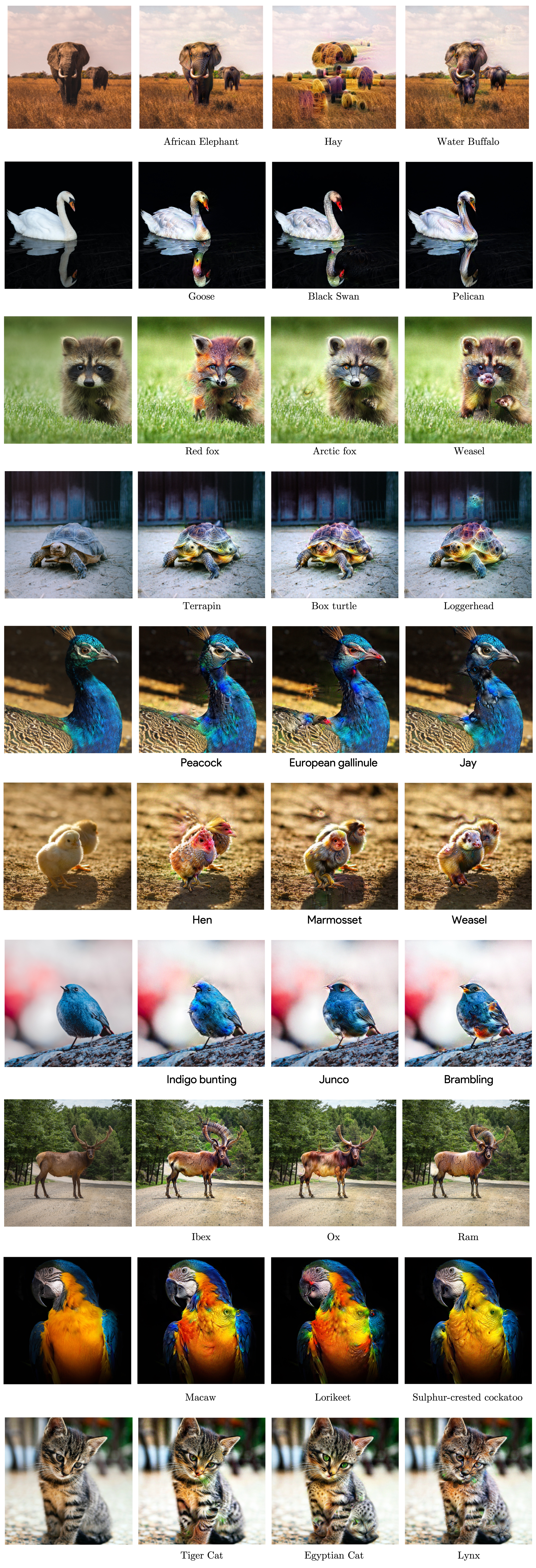}
   \includegraphics[width=0.49\textwidth]{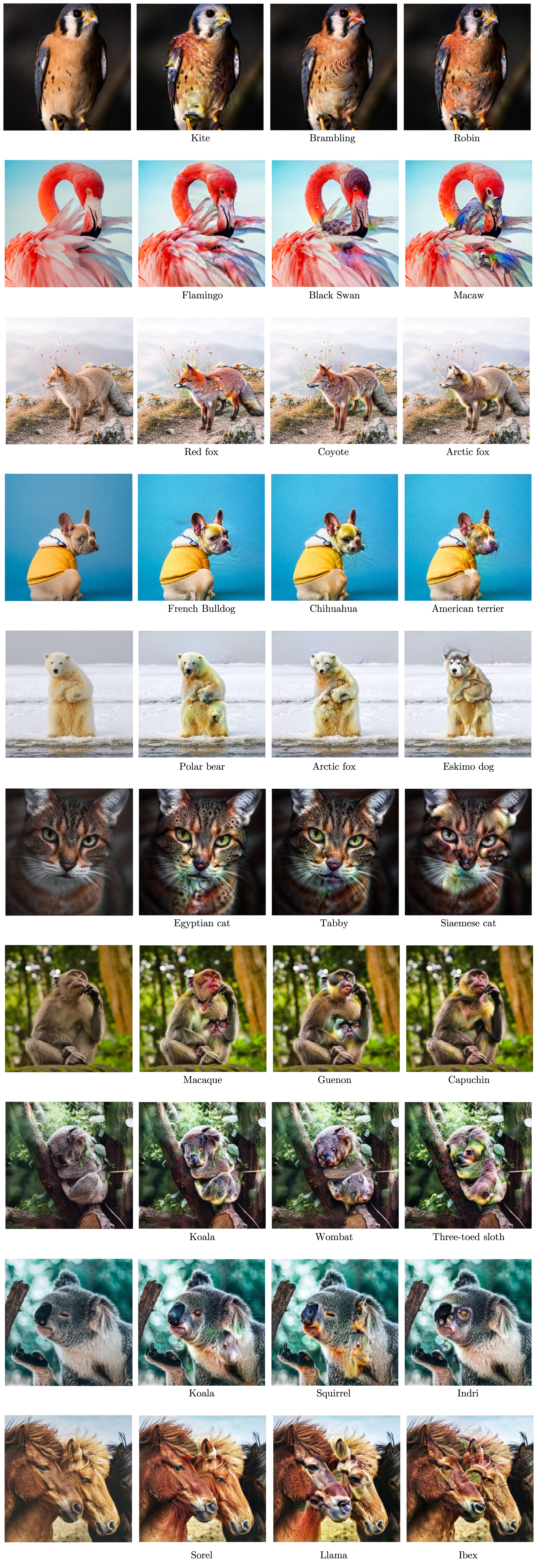}
   \caption{Additional Class-wise accentuations}
\end{figure}

\section{Code}

 Code for this project is available at \href{https://anonymous.4open.science/r/faccent_public-0F00/}{this github repository}.

\section{Human experiment details}\label{sec:human experiment}

For our human evaluation experiment, in each trial we tested people on their ability to correctly identify which of two images would yield a larger activation for a unit in layer mixed4e of InceptionV1. We compared the efficacy of accentuations and maximally activating dataset examples in their ability to improve accuracy on this binary choice task. We additionally included a control condition with no explanation.\par
\paragraph{Experimental dataset generation} We randomly selected 50 units (basis directions) for this experiment and used the following procedure to generate the 3 experimental condition trials for each unit. First, we recorded activations for the unit in response to the Imagenet validation set. We chose the dataset explanation to be that image which resulted in the largest activation, cropped to the activation's effective receptive field\cite{recep_field_1}. Next we chose a target image, i.e the image we are endeavoring to explain. We used a receptive field crop corresponding to an activation that was within the top-1000 largest activations recorded (out of 50000 images * 7 x 7 activation map = 2,450,000 activations), and most cosine dissimilar in the mixed4e channel dimension from the previously selected dataset explanation among this subset.\par
To generate our accentuation explanation, we initialized our optimization at the target image, then optimized using our feature accentuation loss (def \ref{def: fa}), maximizing the activation in mixed4e at the position corresponding to the target image crop, and minimizing the distance in mixed3a across the entire activation map. For our augmentation scheme, we use a more modest minimum crop size of .6, as is necessary for these 'neuron-wise' accentuations, as well as uniform and gaussian noise. We set our regularization \(\lambda =.1\) for all accentuations in the experiment, then optimized for 100 steps using the Adam optimizer \cite{kingma2017adam}, with parameters \(lr = .05\, \;\;\  \beta_{1}=0.9\, \;\;\ \beta_{2}=0.999\).\par 
Finally, to specify the two images participants must choose between for each trial, we wanted two images which were similar to the target image, for which one activated the target unit (the correct choice), but the other did not. To do this, we first specified the incorrect choice as the crop which yielded the largest cosine similarity to the target image in the channel dimension of mixed4e, but nonetheless had 0 activation for the target unit. The correct choice was then chosen to be the crop that yielded the largest activation for the target unit, but had a similar cosine similarity to the target image as the incorrect choice (within \(\pm .05\)).  Throughout, we ensured no two crops within a given trial were taken from the same image. 
\paragraph{Running the Experiment} For each of the 50 units/trials we ran an independent experiment using accentuation explanations, dataset explanations, or no explanation. We ran a total of 298 participants (98 accentuation, 97 dataset, and 98 control) after excluding participants who took outside 3 standard deviations of the average time to complete the experiment and also who had an accuracy outside 3 standard deviations of the average performance by explanation condition. Participants were recruited and paid \$1 for the \textasciitilde 5 minute experiment through Prolific (www.prolific.com). An example trial for each of the 3 explanation types Figure \ref{fig: human exp}.

\begin{figure}
   \centering
   \includegraphics[width=\textwidth]{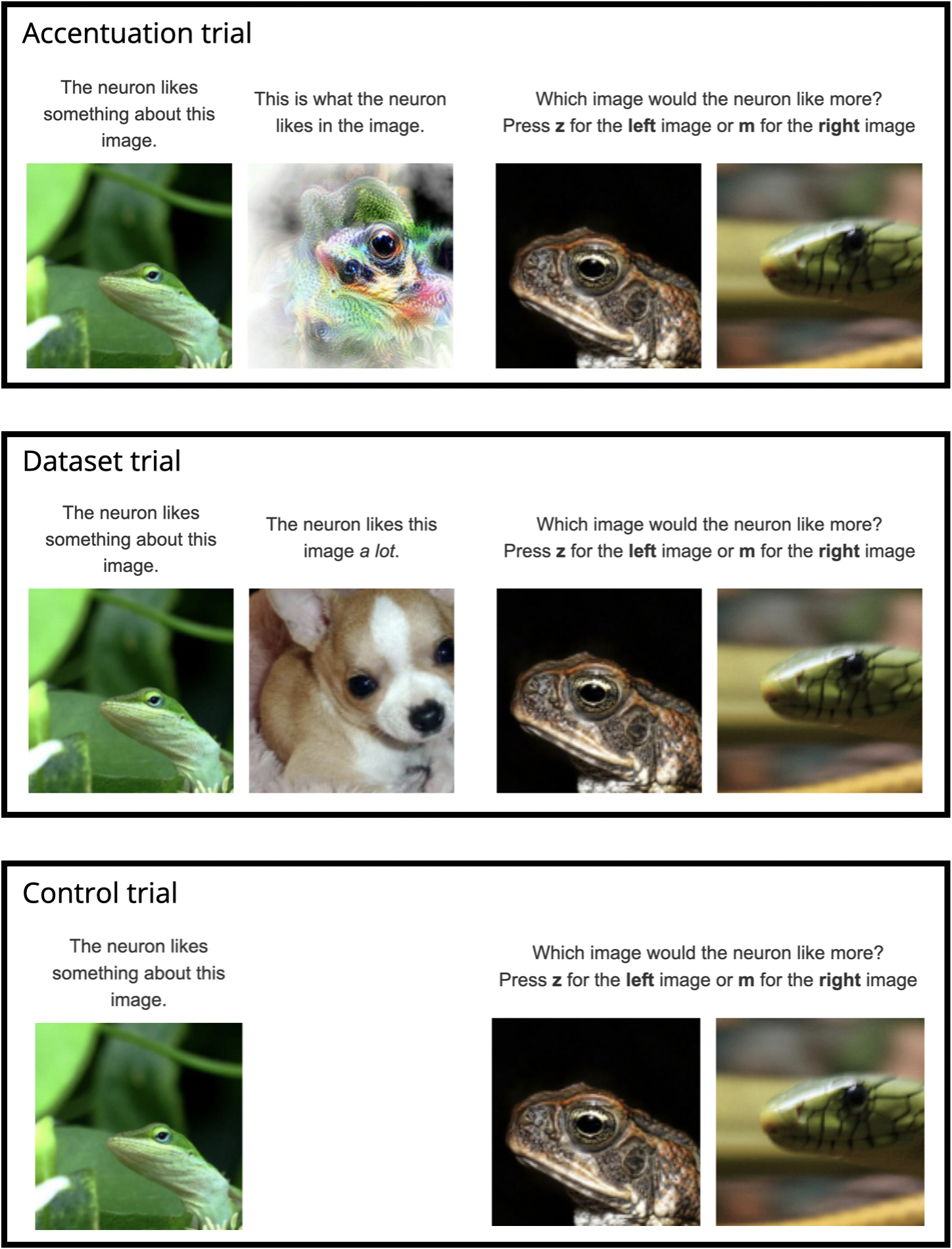}
   \caption{Examples for different explanation trial types in human experiment. The correct answer for this trial is the frog on the left.}
\end{figure}\label{fig: human exp}

\section{Path experiment details}\label{sec:path_app}

For each of the 5 \(\lambda\)s tested (0,.1,.2,.5,10) we generated 1751 accentuations, corresponding to all the correctly predicted images across 50 random classes in the Imagenet \citep{imagenet_cvpr09} validation set. We accentuated each image toward its class label for 100 optimization steps using the Adam optimizer, with a \(.05\) learning rate. We used 16 augmentations each optimizition step, cropping with a \texttt{maximum} box size of .99, and a \texttt{minimum} box size of .05. As an additional augmentation we add gaussian and uniform noise with \(\sigma=.02\) (given initial images normalized to the range [0,1]) each iteration. We regularized through layer \textit{mixed3a}, but note that the optimal regularization layer seems to interact with the layer of \(\pred_{\vv}\). Besides this and \(\lambda\), we observe the above hyperparameters produce quality accentuations in the general case (see section \ref{sec:other models}).\par
Within a given class, we get the average pair-wise correlation for each natural image, and similarly the average correlation of each accentuation to the natural images (excluding the correlations between accentuations and their natural image pair, which would be trivially high). 
\begin{wrapfigure}[38]{l}{0.43\textwidth}
  \centering
  \includegraphics[width=0.4\textwidth]{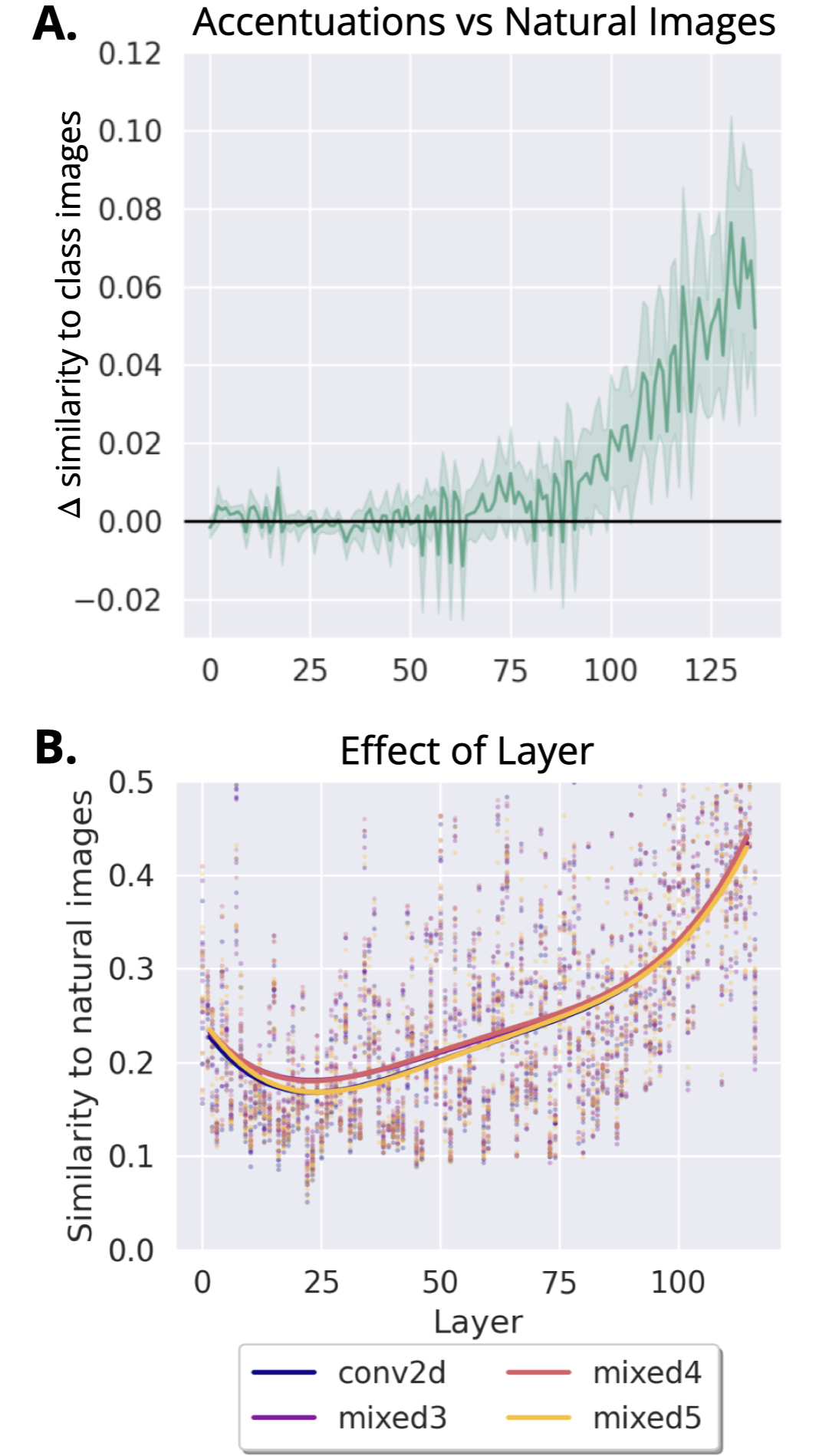}
  \caption{\textbf{A.} The path similarity of feature accentuations to natural images, normalized such that the zero-line corresponds to the path similarity of natural images to themselves. Banded region corresponds to the +/-1 standard deviation across 50 classes tested. \textbf{B.} Small effect of regularization layer on path coherence.)}
  \label{fig:effect layer path}
\end{wrapfigure}
The curves depicted in Figure \ref{fig:path_experiment} are a spline interpolation (degree 2), across the underlying data-points (class-wise average correlations) averaged into 10 bins. We took this approach so as to convey the raw data and the general trend simultaneously. A difficulty with plotting this data stems from the fact that there is a large amount of variance in the correlation measure across nearby layers of different architectural type. We can filter out this variance in Figure \ref{fig:path_experiment}.a and isolate the effect we are interested in by plotting the difference between the correlation measure for accentuations and natural images in each layer. This version of the plot can be seen in Figure \ref{fig:effect layer path}.a. It requires no smoothing, but does not convey the absolute magnitude of the correlation across layers. \par 
In addition to our experiments with \(\lambda\), we tested the effect of regularization layer on the class-wise path similarity metric. We found only a small effect, with the earliest and latest layers tested (\textit{conv2d0} and \textit{mixed5a}) performing slightly worse, in agreement with our qualitative evaluation of the corresponding accentuations.\par
\begin{figure}
   \centering
   \includegraphics[width=\textwidth]{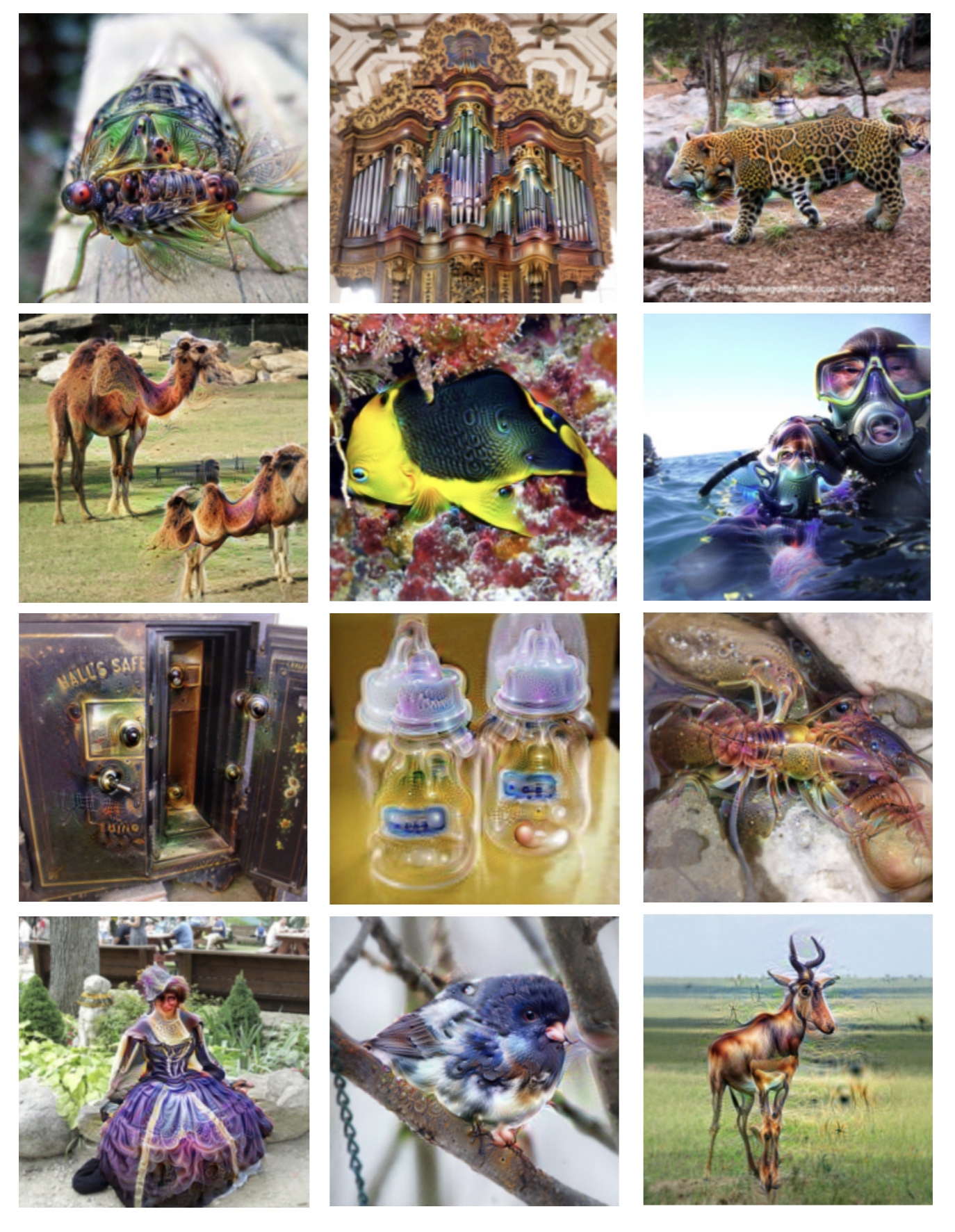}
   \caption{A sample of \textit{'super-natural'} class images, which follow a prototypical path through the model (see section \ref{sec:experiments}). }
   \label{fig:super-natural}
\end{figure}

Figure \ref{fig:super-natural} shows a random sample of our \textit{super-natural} accentuations. These images are processed by way of hidden vectors that better correlate with those for natural class images than the natural images correlate with themselves.

\section{The Proper Parameterization}\label{sec: parameterization}

\begin{figure}[h]
\begin{center}
  \includegraphics[width=\textwidth]{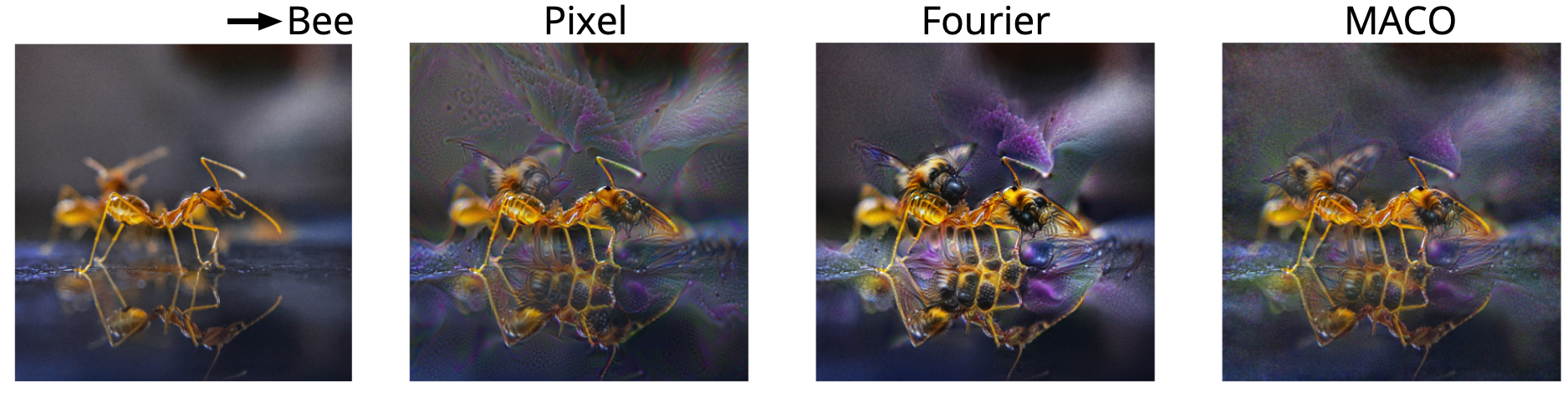}
\end{center}
\caption{\textbf{Effect of image parameterization.}
Modern parameterizations such as Fourier and MACO method enable a more effective management of high frequencies, thereby generating meaningful perturbations. For the remainder of this article, we choose the Fourier parameterization for the InceptionV1 model, although MACO can also be employed, albeit with greater constraints and suitability for deeper models.
}\label{fig:param_qual}
\end{figure}

We proceed with an examination of the quality of the images generated by \FA~  employing distinct image parameterizations. Our experimentation centers on the InceptionV1 network, which is well-established in the feature visualization literature.
In this evaluation, we contrast three parameterizations of increasing complexity: \textit{pixel}, \textit{Fourier} \citep{olah2017feature,mordvintsev2018differentiable}, and \textit{MACO}, introduced in \cite{fel2023maco}.

The \textit{Fourier} parameterization~\citep{mordvintsev2018differentiable,olah2017feature} is a well-recognized selection for feature visualization. It employs weight factors $\bm{w}$ to emphasize low frequencies and mitigate the optimization of high-frequency (adversarial) noise. This parameterization can be expressed as $\vx = \Finv(\vz \odot \bm{w})$, where $\odot$ denotes the Hadamard product.
The \textit{MACO} parameterization, akin to \textit{Fourier}, operates by representing the image in frequency domain, but in polar form $\vz = \bm{r} e^{i \bm{\varphi}}$, thus allowing optimizing only the phase, $\bm{\varphi}$, while preserving the fixed magnitude spectrum to avoid introducing frequencies absent in natural images. This strict constraint on high frequencies contributes to the generation of more natural images, particularly in deep models. In the original work, $\bm{r}$ was set to match the average magnitude spectrum of the ImageNet training set~\citep{imagenet_cvpr09}. However, when applying \textit{MACO} to feature accentuation, we initialize optimization with a target image $\vx_{0}$, thereby constraining $\bm{r}$ to the magnitude spectrum of $\vx_0$ and varying it for each image.
Figure \ref{fig:param_qual} presents an example of feature accentuation using the three different parameterizations. Notably, we observe that the \textit{pixel} parameterization tends to optimize for adversarial noise, as previously noted in the literature. In contrast, the alternative \textit{Fourier} and \textit{MACO} parameterizations introduce perceptible and semantically meaningful changes to the image. However, we find that the \textit{MACO} parameterization produces slightly noisier and lower contrast visualizations; we believe in this context \textit{MACO} is over-constrained. We will therefore use the Fourier parameterization for the rest of the article (see Appendix for more results).

\section{The problem with pixel space regularization}\label{sec:pixel space app}
\begin{figure}
  \centering
  \includegraphics[width=\textwidth]{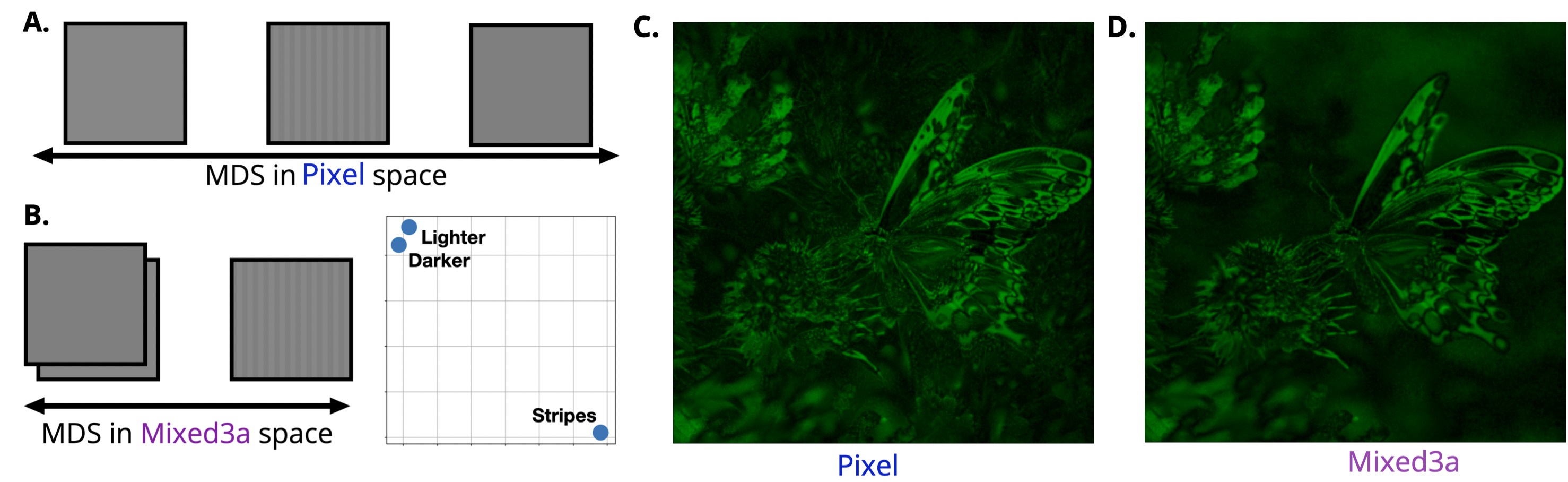}
  \caption{\textbf{A.} Given three grayscale images, one of a homogenous darker shade, one homogenously lighter, and one that iterates between the two shades in a stripe pattern, pixel space difference must put the stripe image between the homogenous images. \textbf{B.} Both perceptually and in the early layers of the model, the striped pattern, a second-ordered statistic, makes it representationally dissimilar to the homogenous images. \textbf{C.} Difference map for accentuation with pixel space regularization, which adds patterns to the background, and  \textbf{D.} latent space regularization, which does not.}
  \label{fig:stripes_mds}
\end{figure}

In the paper we argue that feature accentuations should be regularized such that the optimized image is close to the seed image in the latent space in an early layer, rather than in pixel space as is done with related techniques. We note that pixel space regularization leads to a visible, low amplitude pattern across the entire image, especially in the background, where as latent space regularization yields changes localized to the relevant objects. There is an intuitive explanation for this phenomena, which we illustrate in Figure \ref{fig:stripes_mds}. Observe that the distance between images in pixel space is insensitive to any higher-order statistics in the images, all that matters is the difference of pixel values at each position. This means, for example, that a striped image containing two slightly different pixel values is closer to homogenous images of either value than the homogenous images are to each other (Figure \ref{fig:stripes_mds}.a). However, perceptually it is clear the striped image is the odd-one-out; our visual system 'boosts' the spatial pixel gradient such that we clearly percieve the stripes, even when the homogenous images are indistinguishable. Empirically, the early operations of a neural network appear to do something similar, as the striped image is far from the homogenous images in early layers (Figure \ref{fig:stripes_mds}.b). Small changes in individual pixel values can introduce large changes in our perception and model representation when they introduce new higher-order statistics, like faint edges. Thus feature accentuations with pixel-space regularization optimizes for faint patterns throughout, as observed in our experiments. Figure \ref{fig:stripes_mds}.c and .d show absolute pixel difference maps from the original image for accentuations generated with pixel space and latent space regularization. Changes in the background are patterned for the pixel regularized image, but smooth for the latent regularized one.

\section{masking}
Here we will briefly expound upon our normalized attribution method, put forward in section \ref{sec:masking}. For a data sample \(\bm{X}\) we consider a set of attribution maps  $\{\explainer(\pred_{\vv}, \vx)\} \; | \; \vx \in \bm{X}\}$. Let \(\bm{v}\) represent the flattened, ordered vector of every scalar attribution in this set. We can then choose a percentile range \((p_{1}, \; p_{2})\), such that values outside the range are fully masked/fully visible in the upsampled mask, and values in between are partially masked. We can then normalize every element, \(u\), of a particular attribution map, \(\explainer(\pred_{\vv}, \vx)\), by;
\begin{equation}
\mathcal{N}(u;\bm{v},p_{1},p_{2}) = 
\begin{cases} 
0 & \text{if } u \leq P(\bm{v}, p_1) \\
\frac{u - P(\bm{v}, p_1)}{P(\bm{v}, p_2) - P(\bm{v}, p_1)} & \text{if } P(\bm{v}, p_1) < u < P(\bm{v}, p_2) \\
1 & \text{if } u \geq P(\bm{v}, p_2)
\end{cases}
\end{equation}

By setting \(p_{1}\), the user can control how globally salient a region must be before it constitutes a partial expression of the feature and can be unmasked. Conversely, by setting \(p_{2}\), the user determines the percentile past which a feature is 'fully expressed', and fully unmasked. As previously stated, this approach is agnostic to the attribution method used; for example we applied this normalization approach to \textit{gradCAM} \citep{GradCAM}  attribution maps to generate the masks for Figure \ref{fig:misclassification}, but simply used the features activation map itself as our attribution for latent accents in Figure \ref{fig:latents}, as convolutional layers are already spatialized.\par

%

\section{Randomization Sanity Check}
\begin{figure}
  \centering
  \includegraphics[width=0.95\textwidth]{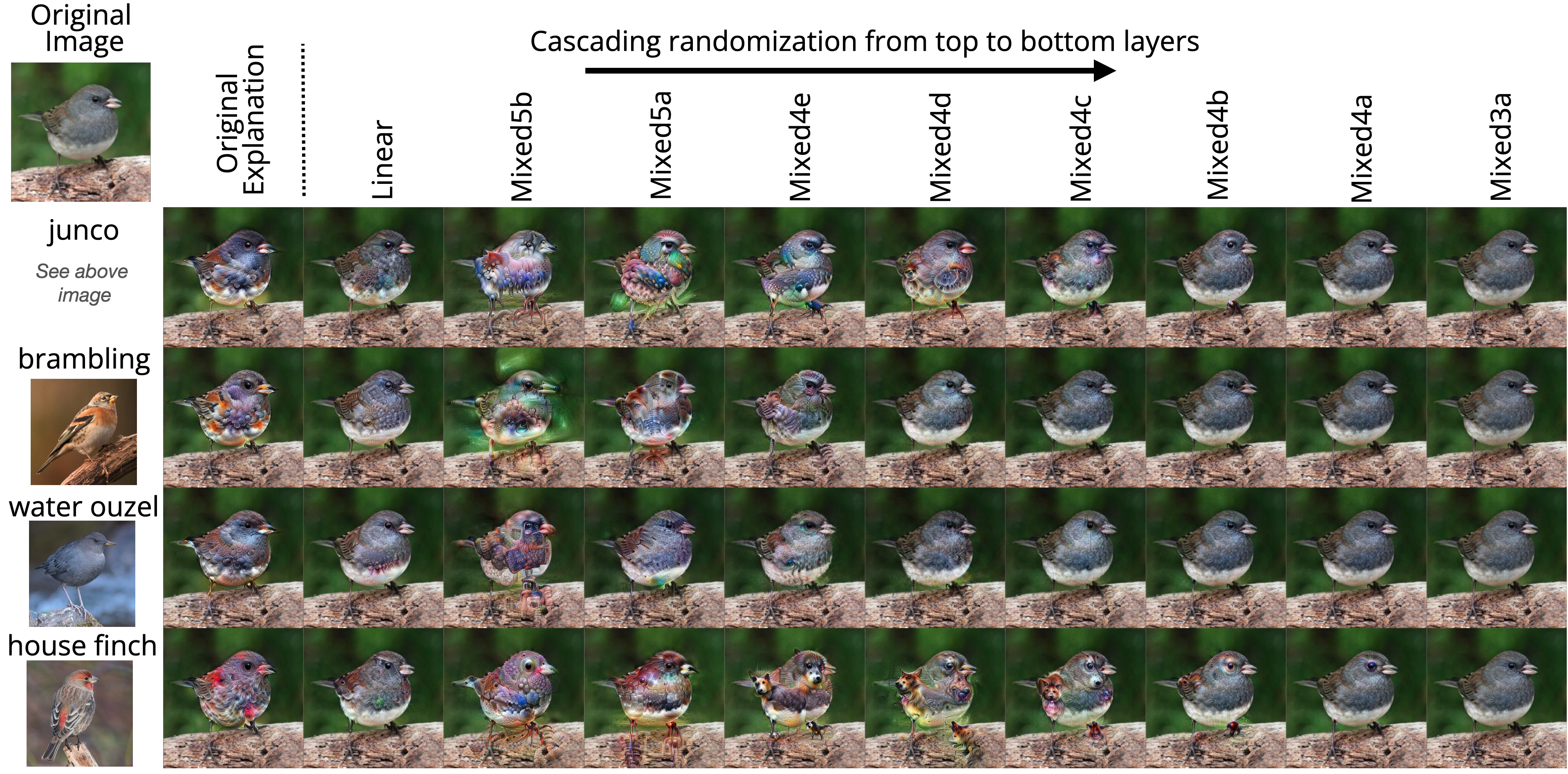}
  \caption{The 'cascading randomization' sanity check from \citep{sanity-checks} performed on feature accentuation. From left to right: feature accentuation is performed on a model with more and more layers randomized. From top to bottom: different logit targets (corresponding to different bird classes) are used as the accentuation target. }
  \label{fig:randomization}
\end{figure}

\cite{sanity-checks} suggest a sanity check that any good explainability method should pass; an explanation for logit \(\bm{y}\) of image \(\bm{x}\) should perceptibly change when the network weights are randomized. This is because the function that returns \(\bm{y}\) from \(\bm{x}\), \(f(\bm{x}) = \bm{y}\), is very different after weight randomization, thus it should require a different explanation. The authors show several attribution techniques fail this sanity check, here we check how feature accentuation fairs. Figure \ref{fig:randomization} shows feature accentuations under 'cascading randomization', where an increasing number of layers are randomization from top to bottom. We use the same 'junco' image used in the intial work, but also accentuate towards 4 different logits, those with the highest activation to the junco image (in the unrandomized model). We observe that while the original feature visualizations change the image towards the targeted bird class, randomizing layers pushes the visualization towards seemingly random tagets. We use a \(\lambda\) 0f .1 for every visualization, and observe that as additional layers are randomized the regularization term tends to dominate and the accentuations stop altering the image at all.

\clearpage

\section{The 'sensitive region' for \texorpdfstring{$\lambda$}{lambda}}\label{sec:regularization_extra}

While qualitatively evaluating the effects of regularization strength \(\lambda\), we observe that for a given image \(\bm{x}\) and feature target \(\pred_{\vv}\), proper accentuation requires \(\lambda\) be within a certain 'sensitive region'. By this we mean, as lambda increases the resultant accentuation eventually becomes perceptually indistinguishable from the seed image, and conversely as \(\lambda\) exponentially decays towards zero, the resultant visualizations becomes indistinguishable from the unregularized accentuation. Increasing \(lambda\) within the intermediate region leads to visualizations that perceptually interpolate between the unregularized accentuation and the seed image. Here we consider how this sensitive region can be identified automatically, without visual inspection. \par
\begin{figure}[t!]
  \centering
  \includegraphics[width=0.95\textwidth]{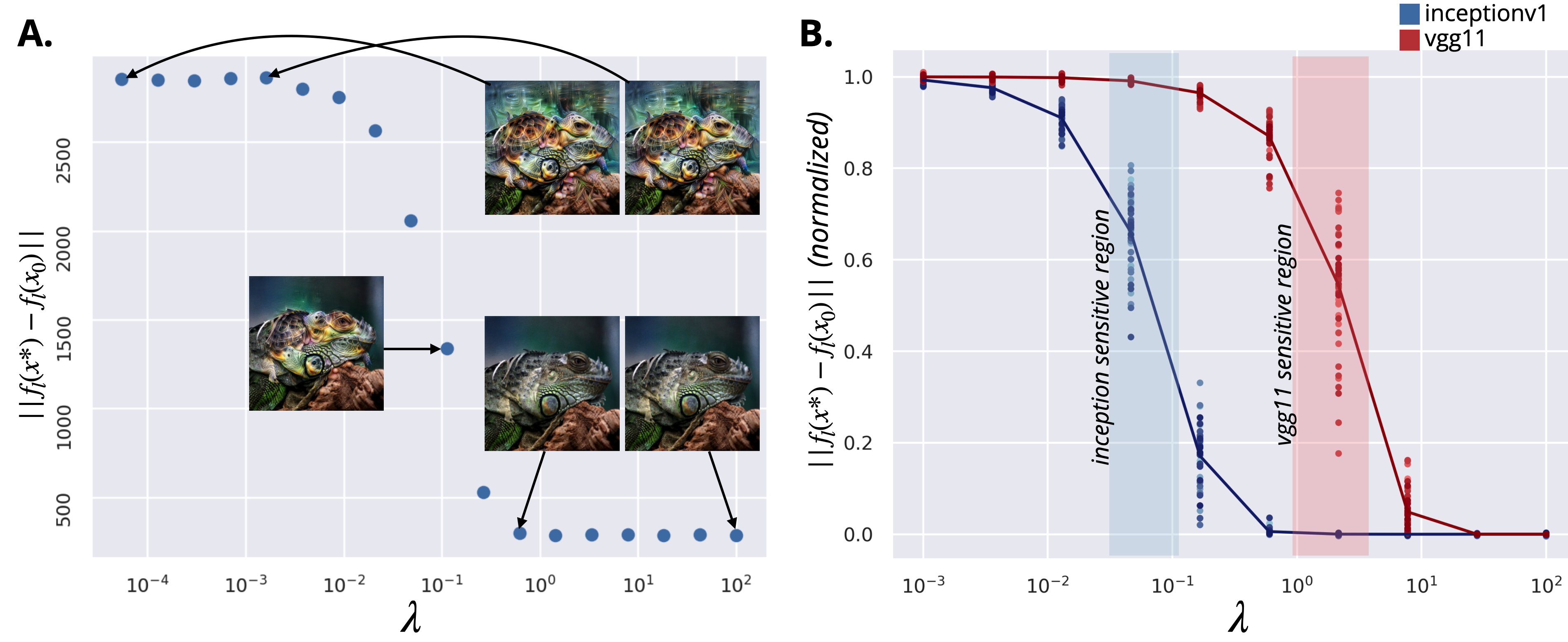}
  \caption{In \textbf{A.} we show the correspondence between the appearance of an accentuation at different \(\lambda\) values and the distance of the accentuation from the seed image in the regularization layer. We note a \textit{sensitive region} for \(lambda\) in which changes perceptibly alter the level of accentuation. In \textbf{B.} we observe that the location of this sensitive region differs significantly across models, but not across images/features in a given model layer.}
  \label{fig:sensitive region}
\end{figure}

 
Such automation requires we establish a simple metric, measurable in the model, that corresponds with the perceptual changes induced by changing \(\lambda\). The simplest candidate is the regularization term itself, \(|| \predl (\vx^{*}) - \predl(\vx_0) ||\), that is, the distance between the representations of the accentuated image and the seed image in the regularization layer. In Figure \ref{fig:sensitive region}.a we show for a single accentuation (iguana -> terrapin (turtle)) how changes in this distance correspond to perceptual change. As \(\lambda\) becomes large/small this distance becomes fixed, as does 
the appearance of the accentuated image. This suggests the sensitive region can be found by first measuring the bounding distances for the accentuation from the seed image (at very small and very large \(\lambda\)), then searching for a \(\lambda\) that yields a distance between these bounds. \par
It would be time consuming to search for \(\lambda\) each time we want to run a new accentuation, but fortunately we find that for a given model layer the sensitive region is consistent across images/features. This can be seen in Figure \ref{fig:sensitive region}.b; here we pass 50 random images through InceptionV1 and VGG11 \citep{vgg}, and accentuate each image towards its second highest logit across a range of regularization strengths. We plot \(\lambda\) on the x-axis as before, and on the y-axis we plot \textit{min-max normalized} the accentuation's distance from the seed image, where the minimums and maximums correspond to the maximum and minimum distances measured at extreme  \(\lambda\)s (\(10^{-6},10^{6}\)). We see that while each model requires a different setting for \(\lambda\), there is a much smaller effect across images. As the sensitive region depends on the model and not the image/feature, for many use cases an appropriate \(\lambda\) must only be found once, and then many different examples can be generated using that setting, as we have done throughout this work.

\clearpage

\section{A note on Learning Rate}
\begin{wrapfigure}[20]{r}{0.45\textwidth}
  \centering
  \includegraphics[width=0.43\textwidth]{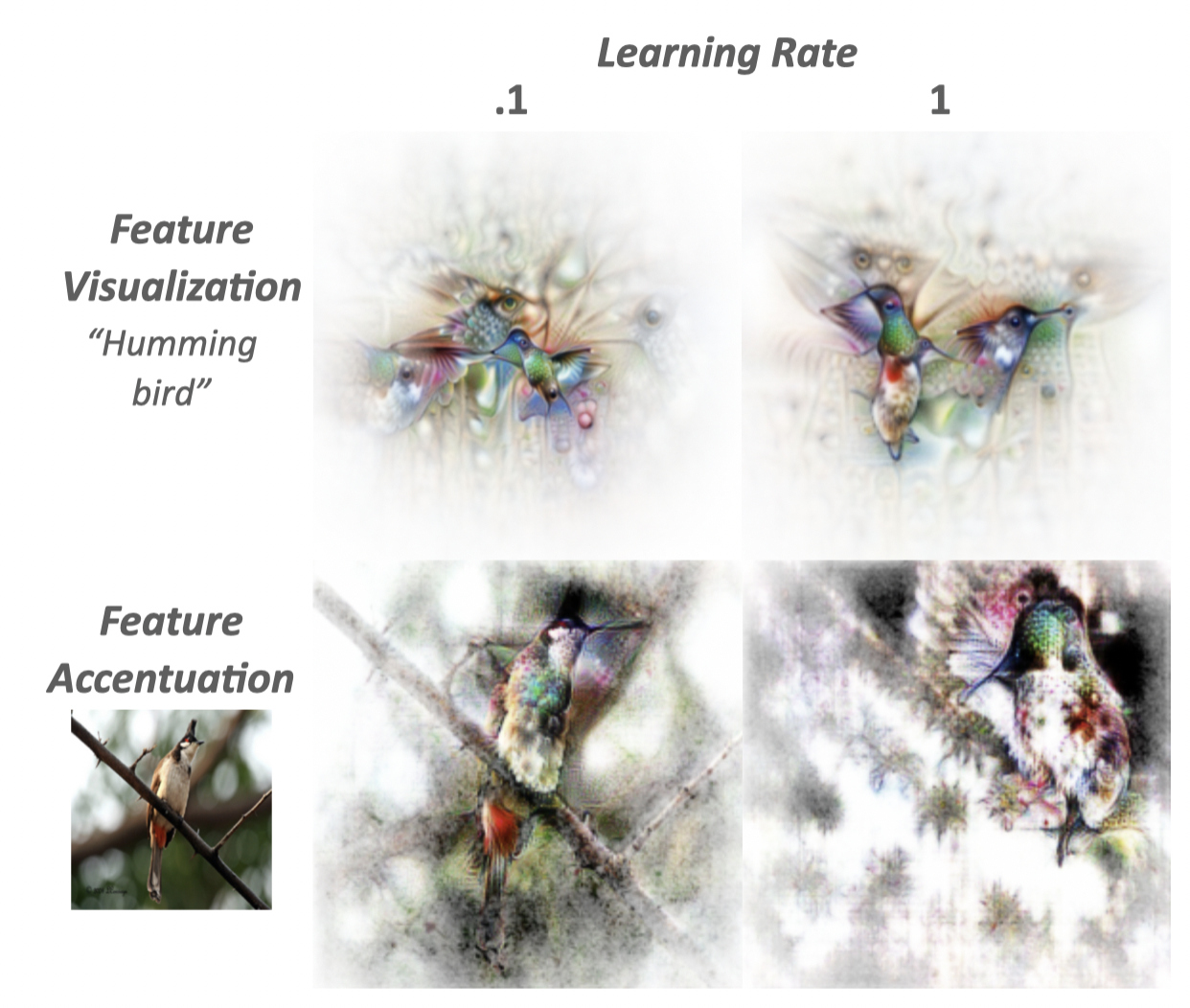}
  \caption{feature visualizations and accentuations for 'hummingbird' at two learning rates. }
  \label{fig:lr}
\vspace{-10mm}
\end{wrapfigure}
We observe learning rate that works for noise-seeded feature visualizations may be too large for feature accentuation, causing the visualization to deviate drastically from the target image in the the initial steps and never make it back. A smaller learning rate keeps the visualization perceptually similar to the seed image (figure \ref{fig:lr}). 

\section{Additional examples for Hyperparameters}

In the interest of conserving space, we initially utilized a single example image to illustrate the impact of manipulating hyperparameters that we consider pivotal for \FA. However, to bolster our demonstration, we present here a series of supplementary examples drawn from our comprehensive testing dataset, aimed at showcasing the consistent and robust nature of the described effects across various instances. We deliberately choose to show some examples multiples times, so the reader can get a sense for how images change along multiple hyperparameter axes.

\begin{figure}[ht!]
  \centering
  \includegraphics[width=0.95\textwidth]{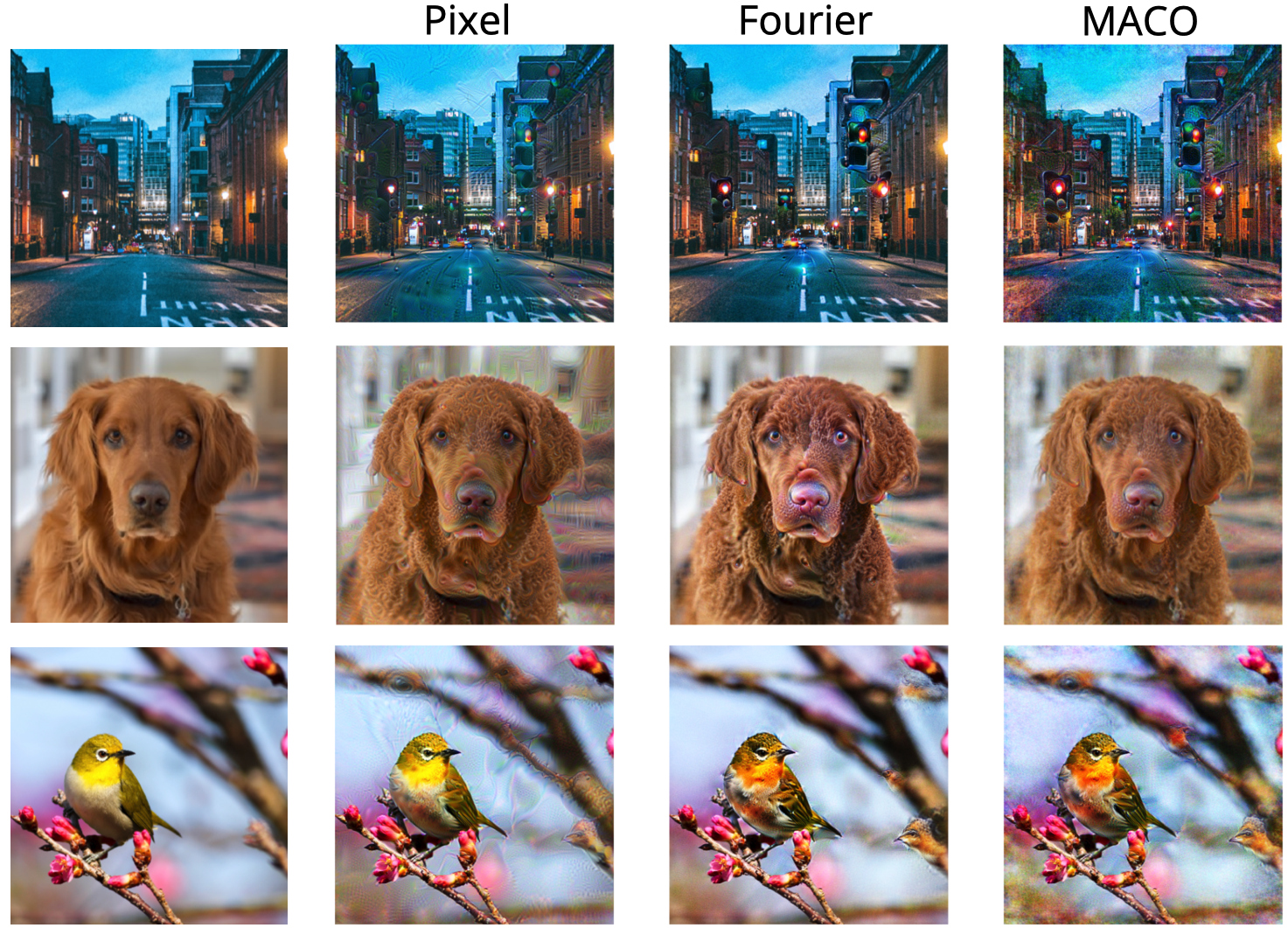}
  \caption{Additional example for our set of image parametrization.}
  \label{fig:other_param}
\end{figure}

\paragraph{Parametrization.} The Fourier and MACO parametrization, while undoubtedly more intricate, emerge as good contenders in generating meaningful perturbations. 
On the other hand, the Pixel parametrization method, while comparatively simpler in its approach, easily introduce adversarial pattern. The seemingly basic alterations to pixel values can yield strikingly impactful results on latent representation, akin to the subtleties found in adversarial perturbations.

\begin{figure}[ht!]
  \centering
  \includegraphics[width=0.95\textwidth]{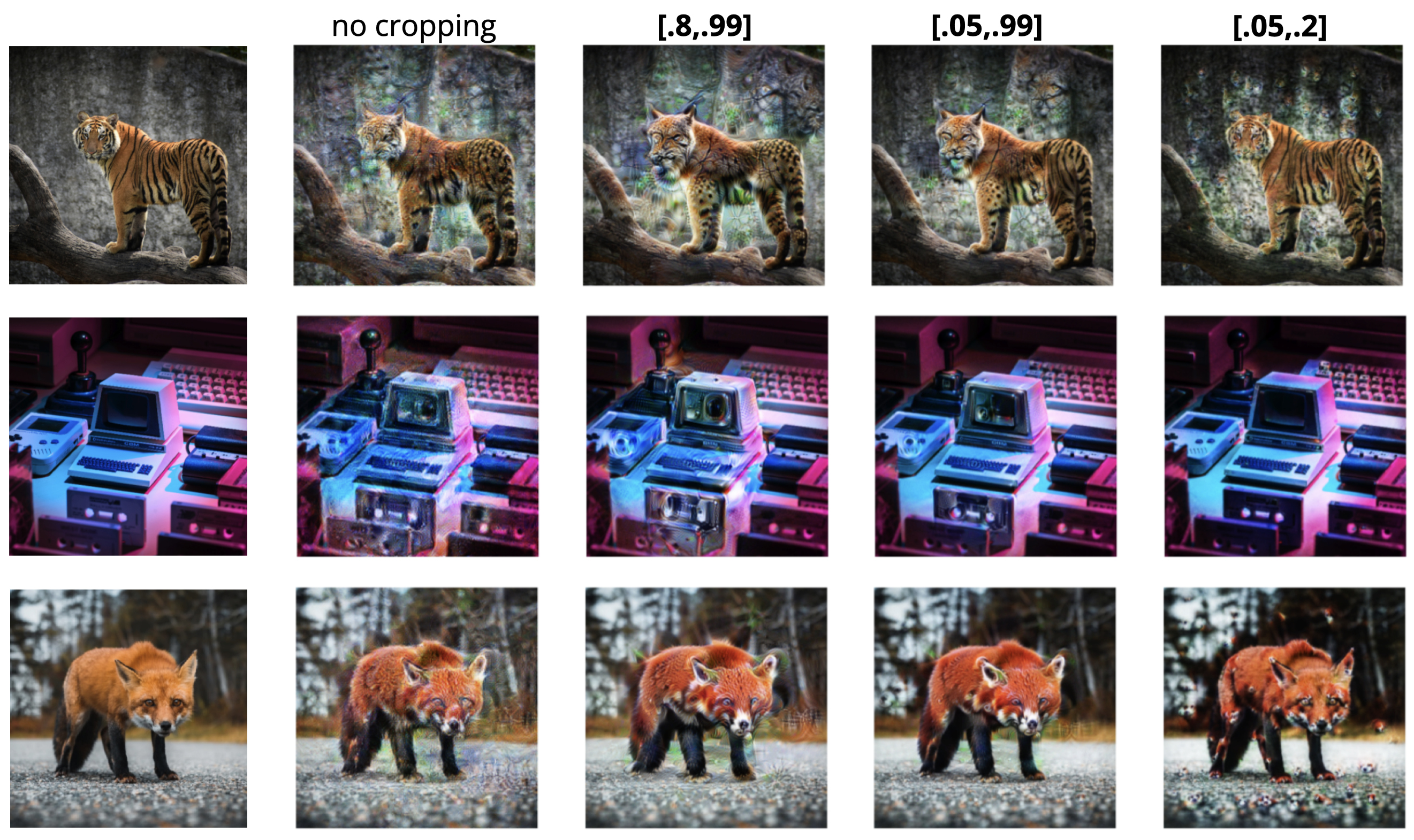}
  \caption{Additional cropping augmentation examples}
  \label{fig:other_crop}
\end{figure}

\paragraph{Augmentation.}
With small crop augmentations, the regularizer and feature detector are essentially \textit{zooming in} and making local edits to the image. Unsurprisely, this yields crisper accentuations. However, small crops can also lead to miniature hallucinatory features scattered throughout the accentuation, especially problematic given we seek explanations for the original, uncropped image. Supplying a uniform mixture of crops each optimization step (as in the fourth column of Figure \ref{fig:other_crop}) seems to regularize against these hallucinations.

\begin{figure}[ht!]
  \centering
  \includegraphics[width=0.95\textwidth]{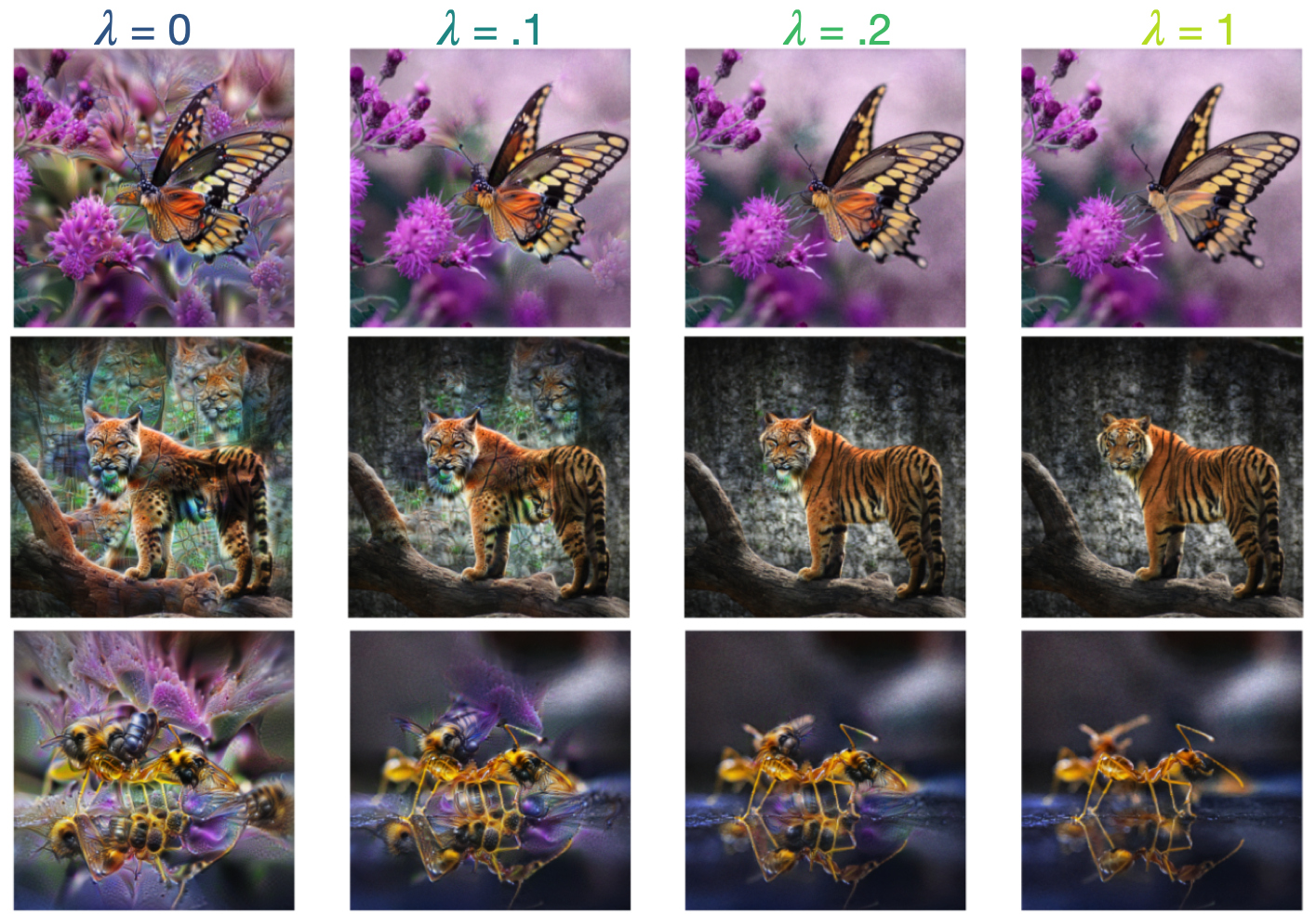}
  \caption{Additional regularization lambda examples}
  \label{fig:other_lambda}
\end{figure}

\begin{figure}[ht!]
  \centering
  \includegraphics[width=0.95\textwidth]{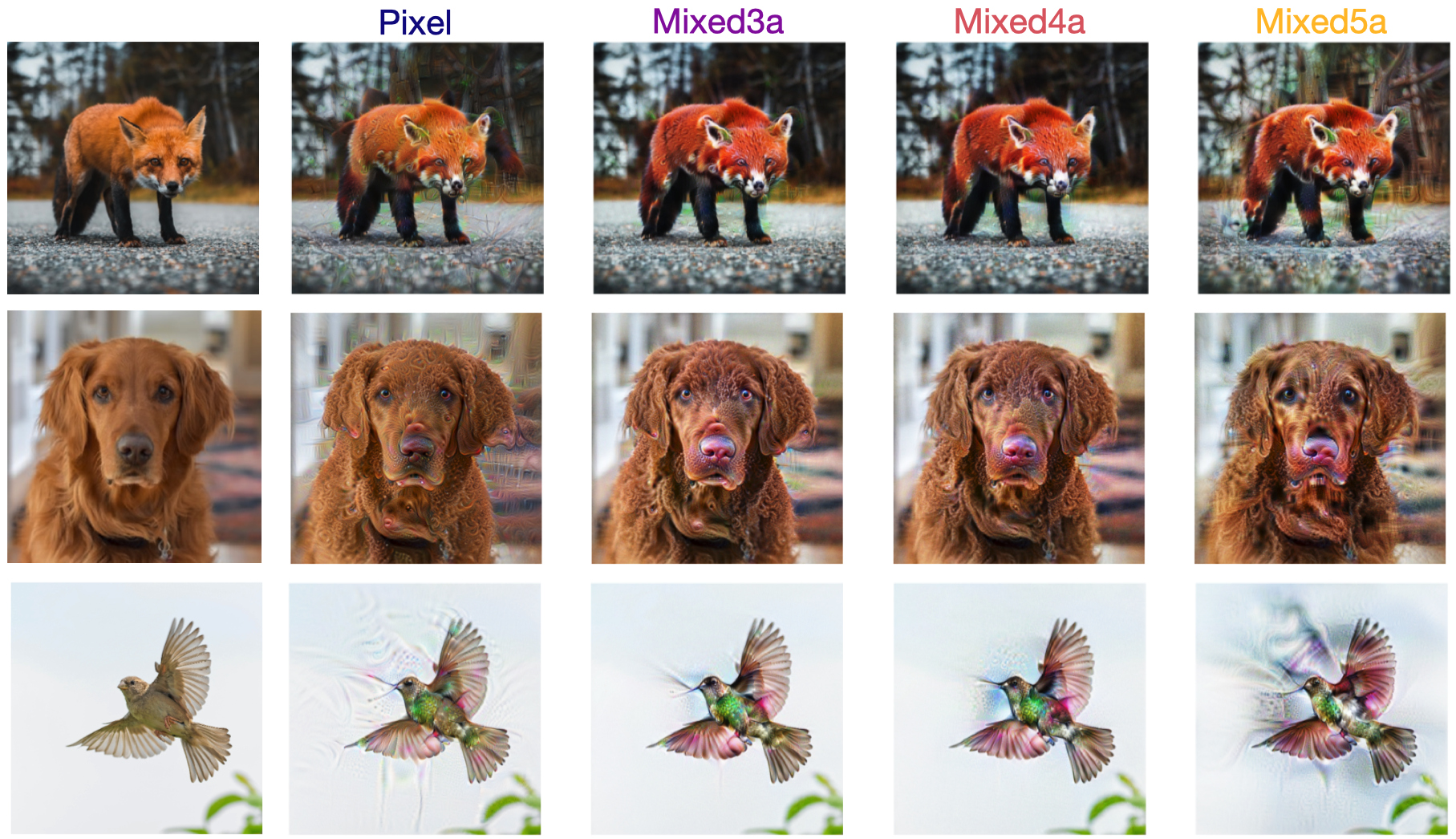}
  \caption{Additional regularization layer examples}
  \label{fig:other_layer}
\end{figure}

\begin{figure}[ht!]
  \centering
  \includegraphics[width=0.95\textwidth]{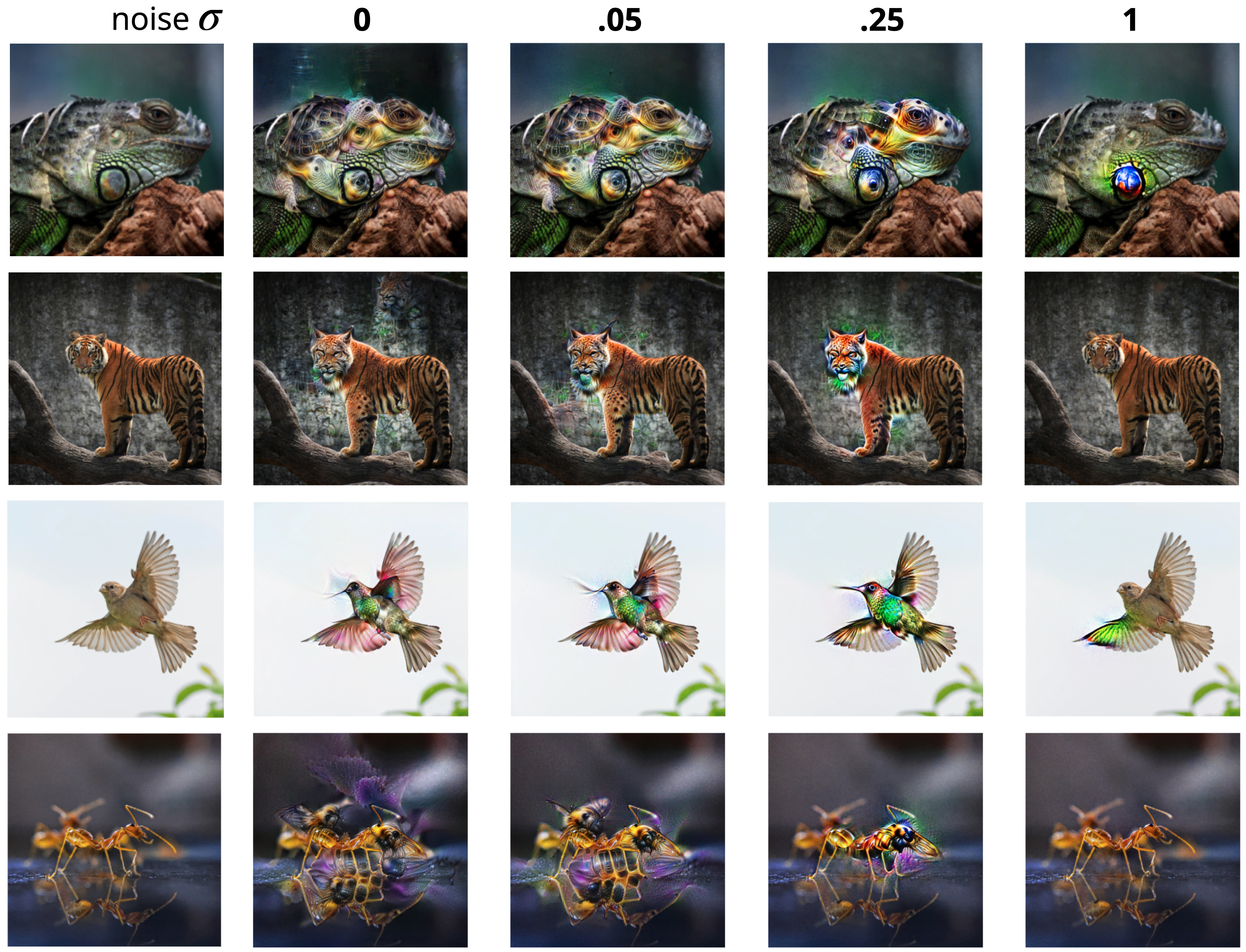}
  \caption{The effects of gaussian and uniform pixel noise as part of the augmentations \(\bm{\tau}\), given noise with different standard deviations \(\sigma\). }
  \label{fig:noise_effect}
\end{figure}

\paragraph{Regularization.}
One of the major hyperparameter influencing our optimization process is $\lambda$ which serves as a regularizer. It plays a vital role in striking a balance between accentuation and preservation of the original data characteristics. More examples are shown confirming that a lower $\lambda$ value can result in more pronounced accentuations, while a higher value tends to maintain the fidelity of the input data. We find the influence of \(\lambda\) interacts with some other hyperparameters discussed here, namely the regularization layer and image parameterization. Given this, to facilitate a fair comparison between different parameterizations/regularization layers we generated accentuations at \(\lambda = 0.05,.1,.5,1,5,10\) for each, then chose the most natural looking image.

\paragraph{Effect of the Layer}
Examining the impact of layer selection within our neural network architecture, we provide additional examples showing the effect of this hyperparameter on \FA.

The choice of the layer has a notable influence on the perturbation:  different layers within our neural network exhibit distinct tendencies in accentuating specific features or patterns within the input data. Early layer tends to preserve pixel information while latter seems to allow greater perturbation but preserve semantic (and often class) information.

\paragraph{Noise}

To increase robustness of feature visualizations, \cite{fel2023maco} uses the addition of gaussian and uniform pixel noise as part of their augmentation scheme. The effects of such noise can be seen in Fig \ref{fig:noise_effect}. We find that feature accentuations work well with no noise, thus we left this augmentation for the appendix. We find that when used in conjuction with or approach to regularization, adding noise tends to focus the areas in which feature accentuation augments the image. When a large amount of noise is applied the regularization term fully dominates and no perceptible changes are made to the image.

\section{Feature Accentuations on other models}\label{sec:other models}

\begin{figure}[ht!]
  \centering
  \includegraphics[width=0.95\textwidth]{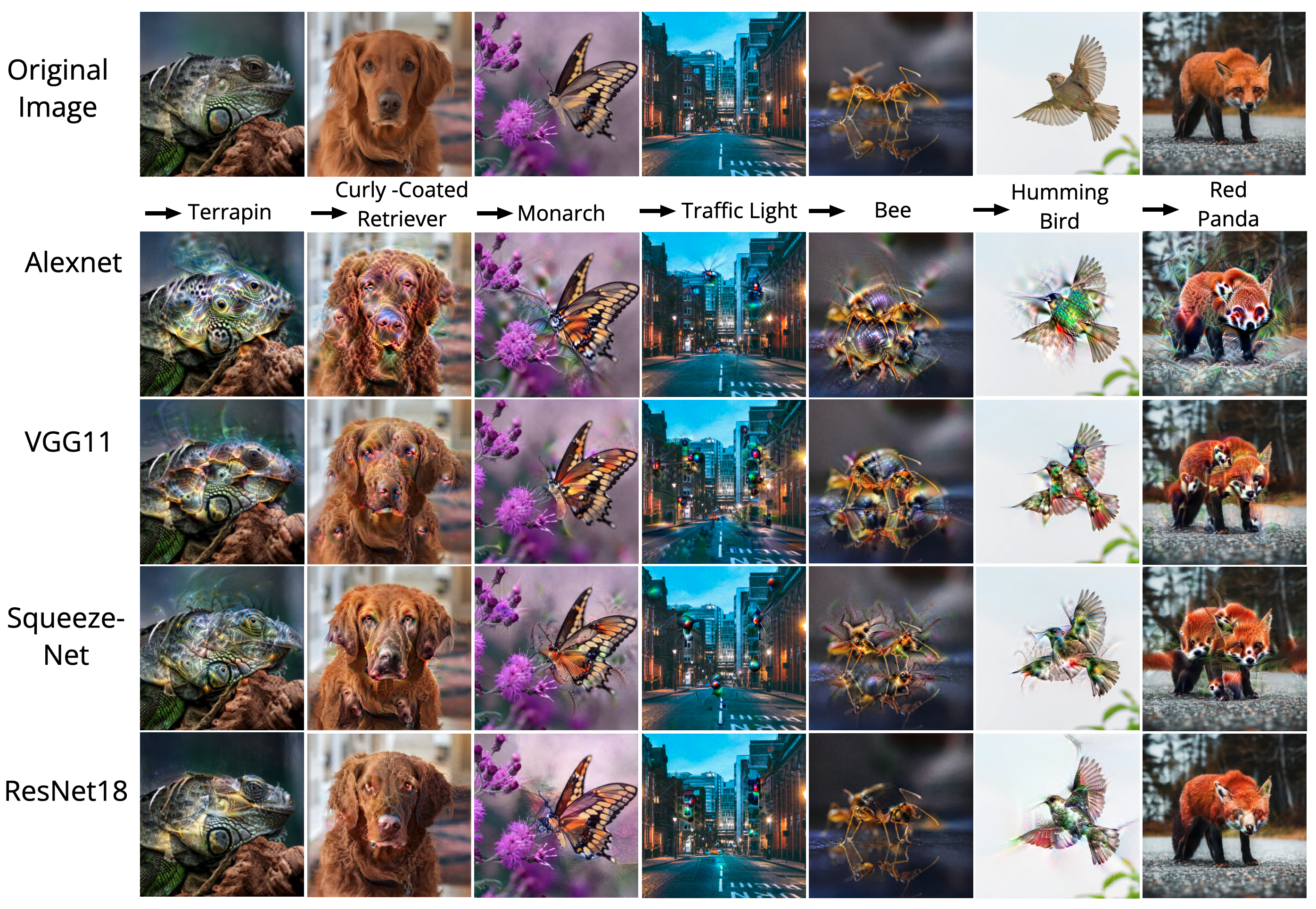}
  \caption{Feature accentuations towards logits for other popular Imagenet trained models. }
  \label{fig:other_models}
\end{figure}

Throughout this work we have focused on the InceptionV1 \citep{mordvintsev2015inceptionism} model given its prevalence in the feature visualization literature, but we find the general recipe put forth in this work is effective for other models as well. Fig \ref{fig:other_models} shows the effects of accentuations on several image/logit pairs for Alexnet \citep{alexnet}, VGG11 \citep{vgg}, SqueezeNet \citep{squeezenet}, and ResNet18 \citep{resnet} models. Regularization \(\lambda\) was set individually per model, to the 'sensitive region' as described in section \ref{sec:regularization_extra}, and the regularization layer was set to the ReLU layer in each model closest to $1/4$ model depth. All other accentuation hyperparameters are fixed across all examples to those hyperparameters described in section \ref{sec:path_app}.  

\end{document}

%% file: math_commands.tex
\usepackage{amsmath,amsfonts,bm}

\def\eqref#1{equation~\ref{#1}}

\def\1{\bm{1}}

\def\va{{\bm{a}}}

\def\vv{{\bm{v}}}

\def\vx{{\bm{x}}}
\def\vy{{\bm{y}}}
\def\vz{{\bm{z}}}

\DeclareMathAlphabet{\mathsfit}{\encodingdefault}{\sfdefault}{m}{sl}
\SetMathAlphabet{\mathsfit}{bold}{\encodingdefault}{\sfdefault}{bx}{n}

\DeclareMathOperator*{\argmax}{arg\,max}

%% file: main.bbl
\begin{thebibliography}{91}
\providecommand{\natexlab}[1]{#1}
\providecommand{\url}[1]{\texttt{#1}}
\expandafter\ifx\csname urlstyle\endcsname\relax
  \providecommand{\doi}[1]{doi: #1}\else
  \providecommand{\doi}{doi: \begingroup \urlstyle{rm}\Url}\fi

\bibitem[Adebayo et~al.(2018{\natexlab{a}})Adebayo, Gilmer, Muelly, Goodfellow, Hardt, and Kim]{adebayo2018sanity}
Julius Adebayo, Justin Gilmer, Michael Muelly, Ian Goodfellow, Moritz Hardt, and Been Kim.
\newblock Sanity checks for saliency maps.
\newblock In \emph{Advances in Neural Information Processing Systems (NIPS)}, 2018{\natexlab{a}}.

\bibitem[Adebayo et~al.(2018{\natexlab{b}})Adebayo, Gilmer, Muelly, Goodfellow, Hardt, and Kim]{sanity-checks}
Julius Adebayo, Justin Gilmer, Michael Muelly, Ian Goodfellow, Moritz Hardt, and Been Kim.
\newblock Sanity checks for saliency maps.
\newblock \emph{Advances in neural information processing systems}, 31, 2018{\natexlab{b}}.

\bibitem[Akhtar \& Mian(2018)Akhtar and Mian]{akhtar2018threat}
Naveed Akhtar and Ajmal Mian.
\newblock Threat of adversarial attacks on deep learning in computer vision: A survey.
\newblock \emph{Ieee Access}, 6:\penalty0 14410--14430, 2018.

\bibitem[Audun(2015)]{AudunGoogleNet}
M.~Øygard Audun.
\newblock Visualizing googlenet classes.
\newblock \emph{URL: https://www.auduno.com/2015/07/29/visualizing-googlenet-classes/}, 2\penalty0 (3), 2015.

\bibitem[Augustin et~al.(2020)Augustin, Meinke, and Hein]{augustin2020adversarial}
Maximilian Augustin, Alexander Meinke, and Matthias Hein.
\newblock Adversarial robustness on in-and out-distribution improves explainability.
\newblock In \emph{European Conference on Computer Vision}, pp.\  228--245. Springer, 2020.

\bibitem[Augustin et~al.(2022)Augustin, Boreiko, Croce, and Hein]{augustin2022diffusion}
Maximilian Augustin, Valentyn Boreiko, Francesco Croce, and Matthias Hein.
\newblock Diffusion visual counterfactual explanations.
\newblock \emph{Advances in Neural Information Processing Systems}, 35:\penalty0 364--377, 2022.

\bibitem[Bach et~al.(2015)Bach, Binder, Montavon, Klauschen, Müller, and Samek]{bach2015pixel}
Sebastian Bach, Alexander Binder, Grégoire Montavon, Frederick Klauschen, Klaus-Robert Müller, and Wojciech Samek.
\newblock On pixel-wise explanations for non-linear classifier decisions by layer-wise relevance propagation.
\newblock \emph{Public Library of Science (PloS One)}, 2015.

\bibitem[Bae et~al.(2020)Bae, Noh, and Kim]{bae2020rethinking}
Wonho Bae, Junhyug Noh, and Gunhee Kim.
\newblock Rethinking class activation mapping for weakly supervised object localization.
\newblock In \emph{Computer Vision--ECCV 2020: 16th European Conference, Glasgow, UK, August 23--28, 2020, Proceedings, Part XV 16}, pp.\  618--634. Springer, 2020.

\bibitem[Baehrens et~al.(2010)Baehrens, Schroeter, Harmeling, Kawanabe, Hansen, and M{\"u}ller]{baehrens2010explain}
David Baehrens, Timon Schroeter, Stefan Harmeling, Motoaki Kawanabe, Katja Hansen, and Klaus-Robert M{\"u}ller.
\newblock How to explain individual classification decisions.
\newblock \emph{The Journal of Machine Learning Research}, 11:\penalty0 1803--1831, 2010.

\bibitem[Boreiko et~al.(2022)Boreiko, Augustin, Croce, Berens, and Hein]{boreiko2022sparse}
Valentyn Boreiko, Maximilian Augustin, Francesco Croce, Philipp Berens, and Matthias Hein.
\newblock Sparse visual counterfactual explanations in image space.
\newblock In \emph{DAGM German Conference on Pattern Recognition}, pp.\  133--148. Springer, 2022.

\bibitem[Borowski et~al.(2020)Borowski, Zimmermann, Schepers, Geirhos, Wallis, Bethge, and Brendel]{borowski2020exemplary}
Judy Borowski, Roland~S Zimmermann, Judith Schepers, Robert Geirhos, Thomas~SA Wallis, Matthias Bethge, and Wieland Brendel.
\newblock Exemplary natural images explain cnn activations better than state-of-the-art feature visualization.
\newblock \emph{arXiv preprint arXiv:2010.12606}, 2020.

\bibitem[Chattopadhay et~al.(2018)Chattopadhay, Sarkar, Howlader, and Balasubramanian]{chattopadhay2018grad}
Aditya Chattopadhay, Anirban Sarkar, Prantik Howlader, and Vineeth~N Balasubramanian.
\newblock Grad-cam++: Generalized gradient-based visual explanations for deep convolutional networks.
\newblock In \emph{Proceedings of the IEEE/CVF Winter Conference on Applications of Computer Vision (WACV)}, 2018.

\bibitem[Deng et~al.(2009)Deng, Dong, Socher, Li, Li, and Fei-Fei]{imagenet_cvpr09}
J.~Deng, W.~Dong, R.~Socher, L.-J. Li, K.~Li, and L.~Fei-Fei.
\newblock {ImageNet: A Large-Scale Hierarchical Image Database}.
\newblock In \emph{Proceedings of the IEEE Conference on Computer Vision and Pattern Recognition (CVPR)}, 2009.

\bibitem[desai \& Ramaswamy(2020)desai and Ramaswamy]{desai_2020_WACV}
saurabh desai and Harish~Guruprasad Ramaswamy.
\newblock Ablation-cam: Visual explanations for deep convolutional network via gradient-free localization.
\newblock In \emph{Proceedings of the IEEE/CVF Winter Conference on Applications of Computer Vision (WACV)}, March 2020.

\bibitem[Doshi-Velez \& Kim(2017)Doshi-Velez and Kim]{doshivelez2017rigorous}
Finale Doshi-Velez and Been Kim.
\newblock Towards a rigorous science of interpretable machine learning.
\newblock \emph{{A}r{X}iv e-print}, 2017.

\bibitem[Engstrom et~al.(2019)Engstrom, Ilyas, Santurkar, Tsipras, Tran, and Madry]{engstrom2019adversarial}
Logan Engstrom, Andrew Ilyas, Shibani Santurkar, Dimitris Tsipras, Brandon Tran, and Aleksander Madry.
\newblock Adversarial robustness as a prior for learned representations.
\newblock \emph{arXiv preprint arXiv:1906.00945}, 2019.

\bibitem[Fel et~al.(2021)Fel, Cadene, Chalvidal, Cord, Vigouroux, and Serre]{fel2021sobol}
Thomas Fel, Remi Cadene, Mathieu Chalvidal, Matthieu Cord, David Vigouroux, and Thomas Serre.
\newblock Look at the variance! efficient black-box explanations with sobol-based sensitivity analysis.
\newblock In \emph{Advances in Neural Information Processing Systems (NeurIPS)}, 2021.

\bibitem[Fel et~al.(2022)Fel, Picard, Bethune, Boissin, Vigouroux, Colin, Cad{\`e}ne, and Serre]{fel2022craft}
Thomas Fel, Agustin Picard, Louis Bethune, Thibaut Boissin, David Vigouroux, Julien Colin, R{\'e}mi Cad{\`e}ne, and Thomas Serre.
\newblock Craft: Concept recursive activation factorization for explainability.
\newblock \emph{Proceedings of the IEEE Conference on Computer Vision and Pattern Recognition (CVPR)}, 2022.

\bibitem[Fel et~al.(2023{\natexlab{a}})Fel, Boissin, Boutin, Picard, Novello, Colin, Linsley, Rousseau, Cad{\`e}ne, Gardes, et~al.]{fel2023maco}
Thomas Fel, Thibaut Boissin, Victor Boutin, Agustin Picard, Paul Novello, Julien Colin, Drew Linsley, Tom Rousseau, R{\'e}mi Cad{\`e}ne, Laurent Gardes, et~al.
\newblock Unlocking feature visualization for deeper networks with magnitude constrained optimization.
\newblock \emph{arXiv preprint arXiv:2306.06805}, 2023{\natexlab{a}}.

\bibitem[Fel et~al.(2023{\natexlab{b}})Fel, Ducoffe, Vigouroux, Cad{\`e}ne, Capelle, Nicod{\`e}me, and Serre]{fel2023don}
Thomas Fel, M{\'e}lanie Ducoffe, David Vigouroux, R{\'e}mi Cad{\`e}ne, Mikael Capelle, Claire Nicod{\`e}me, and Thomas Serre.
\newblock Don't lie to me! robust and efficient explainability with verified perturbation analysis.
\newblock In \emph{Proceedings of the IEEE/CVF Conference on Computer Vision and Pattern Recognition}, pp.\  16153--16163, 2023{\natexlab{b}}.

\bibitem[Fel et~al.(2023{\natexlab{c}})Fel, Picard, Bethune, Boissin, Vigouroux, Colin, Cad{\`e}ne, and Serre]{fel2023craft}
Thomas Fel, Agustin Picard, Louis Bethune, Thibaut Boissin, David Vigouroux, Julien Colin, R{\'e}mi Cad{\`e}ne, and Thomas Serre.
\newblock Craft: Concept recursive activation factorization for explainability.
\newblock In \emph{Proceedings of the IEEE/CVF Conference on Computer Vision and Pattern Recognition}, pp.\  2711--2721, 2023{\natexlab{c}}.

\bibitem[Fong \& Vedaldi(2017)Fong and Vedaldi]{Fong_2017}
Ruth~C. Fong and Andrea Vedaldi.
\newblock Interpretable explanations of black boxes by meaningful perturbation.
\newblock In \emph{Proceedings of the IEEE International Conference on Computer Vision (ICCV)}, 2017.

\bibitem[Fu et~al.(2020)Fu, Hu, Dong, Guo, Gao, and Li]{fuaxiom}
Ruigang Fu, Qingyong Hu, Xiaohu Dong, Yulan Guo, Yinghui Gao, and Biao Li.
\newblock Axiom-based grad-cam: Towards accurate visualization and explanation of cnns.
\newblock \emph{arXiv preprint arXiv:2008.02312}, 2020.

\bibitem[Gaziv et~al.(2023)Gaziv, Lee, and DiCarlo]{gaziv2023robustified}
Guy Gaziv, Michael~J Lee, and James~J DiCarlo.
\newblock Robustified anns reveal wormholes between human category percepts.
\newblock \emph{arXiv preprint arXiv:2308.06887}, 2023.

\bibitem[Geirhos et~al.(2023)Geirhos, Zimmermann, Bilodeau, Brendel, and Kim]{geirhos2023dont}
Robert Geirhos, Roland~S. Zimmermann, Blair Bilodeau, Wieland Brendel, and Been Kim.
\newblock Don't trust your eyes: on the (un)reliability of feature visualizations, 2023.

\bibitem[Ghalebikesabi et~al.(2021)Ghalebikesabi, Ter-Minassian, DiazOrdaz, and Holmes]{ghalebikesabi2021locality}
Sahra Ghalebikesabi, Lucile Ter-Minassian, Karla DiazOrdaz, and Chris~C Holmes.
\newblock On locality of local explanation models.
\newblock \emph{Advances in Neural Information Processing Systems (NeurIPS)}, 2021.

\bibitem[Ghiasi et~al.(2021)Ghiasi, Kazemi, Reich, Zhu, Goldblum, and Goldstein]{ghiasi2021plug}
Amin Ghiasi, Hamid Kazemi, Steven Reich, Chen Zhu, Micah Goldblum, and Tom Goldstein.
\newblock Plug-in inversion: Model-agnostic inversion for vision with data augmentations.
\newblock 2021.

\bibitem[Ghiasi et~al.(2022)Ghiasi, Kazemi, Borgnia, Reich, Shu, Goldblum, Wilson, and Goldstein]{ghiasi2022vision}
Amin Ghiasi, Hamid Kazemi, Eitan Borgnia, Steven Reich, Manli Shu, Micah Goldblum, Andrew~Gordon Wilson, and Tom Goldstein.
\newblock What do vision transformers learn? a visual exploration.
\newblock \emph{arXiv preprint arXiv:2212.06727}, 2022.

\bibitem[Goodfellow et~al.(2014)Goodfellow, Shlens, and Szegedy]{goodfellow2014explaining}
Ian~J Goodfellow, Jonathon Shlens, and Christian Szegedy.
\newblock Explaining and harnessing adversarial examples.
\newblock \emph{arXiv preprint arXiv:1412.6572}, 2014.

\bibitem[Goyal et~al.(2019)Goyal, Wu, Ernst, Batra, Parikh, and Lee]{goyal2019counterfactual}
Yash Goyal, Ziyan Wu, Jan Ernst, Dhruv Batra, Devi Parikh, and Stefan Lee.
\newblock Counterfactual visual explanations.
\newblock In \emph{International Conference on Machine Learning}, pp.\  2376--2384. PMLR, 2019.

\bibitem[Hase \& Bansal(2020)Hase and Bansal]{hase2020evaluating}
Peter Hase and Mohit Bansal.
\newblock Evaluating explainable ai: Which algorithmic explanations help users predict model behavior?
\newblock In \emph{Proceedings of the Annual Meeting of the Association for Computational Linguistics (ACL)}, 2020.

\bibitem[He et~al.(2016)He, Zhang, Ren, and Sun]{resnet}
Kaiming He, Xiangyu Zhang, Shaoqing Ren, and Jian Sun.
\newblock Deep residual learning for image recognition.
\newblock In \emph{Proceedings of 2016 IEEE Conference on Computer Vision and Pattern Recognition}, pp.\  770--778. IEEE, 2016.
\newblock \doi{10.1109/CVPR.2016.90}.
\newblock URL \url{http://ieeexplore.ieee.org/document/7780459}.

\bibitem[Iandola et~al.(2016)Iandola, Han, Moskewicz, Ashraf, Dally, and Keutzer]{squeezenet}
Forrest~N. Iandola, Song Han, Matthew~W. Moskewicz, Khalid Ashraf, William~J. Dally, and Kurt Keutzer.
\newblock Squeezenet: Alexnet-level accuracy with 50x fewer parameters and <0.5mb model size, 2016.

\bibitem[Jacovi et~al.(2021)Jacovi, Marasovi{\'c}, Miller, and Goldberg]{jacovi2021formalizing}
Alon Jacovi, Ana Marasovi{\'c}, Tim Miller, and Yoav Goldberg.
\newblock Formalizing trust in artificial intelligence: Prerequisites, causes and goals of human trust in ai.
\newblock In \emph{Proceedings of the 2021 ACM conference on fairness, accountability, and transparency}, pp.\  624--635, 2021.

\bibitem[Kaminski(2021)]{kaminski2021right}
Margot~E Kaminski.
\newblock The right to explanation, explained.
\newblock In \emph{Research Handbook on Information Law and Governance}. Edward Elgar Publishing, 2021.

\bibitem[Kiat(2020)]{lucent}
Lim~Swee Kiat.
\newblock Lucent, 2020.
\newblock https://github.com/greentfrapp/lucent.

\bibitem[Kim et~al.(2021)Kim, Choe, Akata, and Oh]{kim2021keep}
Jae~Myung Kim, Junsuk Choe, Zeynep Akata, and Seong~Joon Oh.
\newblock Keep calm and improve visual feature attribution.
\newblock In \emph{Proceedings of the IEEE/CVF International Conference on Computer Vision}, pp.\  8350--8360, 2021.

\bibitem[Kim et~al.(2022)Kim, Meister, Ramaswamy, Fong, and Russakovsky]{kim2021hive}
Sunnie S.~Y. Kim, Nicole Meister, Vikram~V. Ramaswamy, Ruth Fong, and Olga Russakovsky.
\newblock {HIVE}: Evaluating the human interpretability of visual explanations.
\newblock In \emph{Proceedings of the IEEE European Conference on Computer Vision (ECCV)}, 2022.

\bibitem[Kindermans et~al.(2019)Kindermans, Hooker, Adebayo, Alber, Sch{\"u}tt, D{\"a}hne, Erhan, and Kim]{kindermans2019reliability}
Pieter-Jan Kindermans, Sara Hooker, Julius Adebayo, Maximilian Alber, Kristof~T. Sch{\"u}tt, Sven D{\"a}hne, Dumitru Erhan, and Been Kim.
\newblock \emph{The (Un)reliability of Saliency Methods}, pp.\  267--280.
\newblock Springer International Publishing, Cham, 2019.
\newblock ISBN 978-3-030-28954-6.
\newblock \doi{10.1007/978-3-030-28954-6_14}.
\newblock URL \url{https://doi.org/10.1007/978-3-030-28954-6_14}.

\bibitem[Kingma \& Ba(2017)Kingma and Ba]{kingma2017adam}
Diederik~P. Kingma and Jimmy Ba.
\newblock Adam: A method for stochastic optimization, 2017.

\bibitem[Kop(2021)]{kop2021eu}
Mauritz Kop.
\newblock Eu artificial intelligence act: The european approach to ai.
\newblock In \emph{Stanford - Vienna Transatlantic Technology Law Forum, Transatlantic Antitrust and IPR Developments, Stanford University, Issue No. 2/2021. https://law.stanford.edu/publications/eu-artificial-intelligence-act-the-european-approach-to-ai/}. Stanford-Vienna Transatlantic Technology Law Forum, Transatlantic Antitrust~…, 2021.

\bibitem[Krizhevsky et~al.(2012)Krizhevsky, Sutskever, and Hinton]{alexnet}
Alex Krizhevsky, Ilya Sutskever, and Geoffrey Hinton.
\newblock Imagenet classification with deep convolutional neural networks.
\newblock \emph{Neural Information Processing Systems}, 25, 01 2012.
\newblock \doi{10.1145/3065386}.

\bibitem[Le \& Borji(2017)Le and Borji]{recep_field_1}
Hung Le and Ali Borji.
\newblock What are the receptive, effective receptive, and projective fields of neurons in convolutional neural networks?
\newblock \emph{CoRR}, abs/1705.07049, 2017.
\newblock URL \url{http://arxiv.org/abs/1705.07049}.

\bibitem[Lundberg \& Lee(2017)Lundberg and Lee]{lundberg2017unified}
Scott Lundberg and Su-In Lee.
\newblock A unified approach to interpreting model predictions.
\newblock In \emph{Advances in Neural Information Processing Systems (NIPS)}, 2017.

\bibitem[Madry et~al.(2018)Madry, Makelov, Schmidt, Tsipras, and Vladu]{madry2017pgd}
Aleksander Madry, Aleksandar Makelov, Ludwig Schmidt, Dimitris Tsipras, and Adrian Vladu.
\newblock Towards deep learning models resistant to adversarial attacks.
\newblock \emph{Proceedings of the International Conference on Learning Representations (ICLR)}, 2018.

\bibitem[Mahendran \& Vedaldi(2015)Mahendran and Vedaldi]{mahendran2015understanding}
Aravindh Mahendran and Andrea Vedaldi.
\newblock Understanding deep image representations by inverting them.
\newblock In \emph{Proceedings of the IEEE conference on computer vision and pattern recognition}, pp.\  5188--5196, 2015.

\bibitem[Montavon et~al.(2017)Montavon, Lapuschkin, Binder, Samek, and M{\"u}ller]{montavon2017explaining}
Gr{\'e}goire Montavon, Sebastian Lapuschkin, Alexander Binder, Wojciech Samek, and Klaus-Robert M{\"u}ller.
\newblock Explaining nonlinear classification decisions with deep taylor decomposition.
\newblock \emph{Pattern recognition}, 65:\penalty0 211--222, 2017.

\bibitem[Moosavi-Dezfooli et~al.(2016)Moosavi-Dezfooli, Fawzi, and Frossard]{moosavi2016deepfool}
Seyed-Mohsen Moosavi-Dezfooli, Alhussein Fawzi, and Pascal Frossard.
\newblock Deepfool: a simple and accurate method to fool deep neural networks.
\newblock In \emph{Proceedings of the IEEE conference on computer vision and pattern recognition}, pp.\  2574--2582, 2016.

\bibitem[Mordvintsev et~al.(2015)Mordvintsev, Olah, and Tyka]{mordvintsev2015inceptionism}
Alexander Mordvintsev, Christopher Olah, and Mike Tyka.
\newblock Inceptionism: Going deeper into neural networks.
\newblock 2015.
\newblock URL \url{https://research.googleblog.com/2015/06/inceptionism-going-deeper-into-neural.html}.

\bibitem[Mordvintsev et~al.(2018)Mordvintsev, Pezzotti, Schubert, and Olah]{mordvintsev2018differentiable}
Alexander Mordvintsev, Nicola Pezzotti, Ludwig Schubert, and Chris Olah.
\newblock Differentiable image parameterizations.
\newblock \emph{Distill}, 2018.

\bibitem[Nguyen et~al.(2015)Nguyen, Yosinski, and Clune]{nguyen2015deep}
Anh Nguyen, Jason Yosinski, and Jeff Clune.
\newblock Deep neural networks are easily fooled: High confidence predictions for unrecognizable images.
\newblock In \emph{Proceedings of the IEEE conference on computer vision and pattern recognition}, pp.\  427--436, 2015.

\bibitem[Nguyen et~al.(2016{\natexlab{a}})Nguyen, Dosovitskiy, Yosinski, Brox, and Clune]{nguyen2016synthesizing}
Anh Nguyen, Alexey Dosovitskiy, Jason Yosinski, Thomas Brox, and Jeff Clune.
\newblock Synthesizing the preferred inputs for neurons in neural networks via deep generator networks.
\newblock \emph{Advances in neural information processing systems}, 29, 2016{\natexlab{a}}.

\bibitem[Nguyen et~al.(2016{\natexlab{b}})Nguyen, Yosinski, and Clune]{nguyen2016multifaceted}
Anh Nguyen, Jason Yosinski, and Jeff Clune.
\newblock Multifaceted feature visualization: Uncovering the different types of features learned by each neuron in deep neural networks.
\newblock \emph{Visualization for Deep Learning workshop, Proceedings of the International Conference on Machine Learning (ICML)}, 2016{\natexlab{b}}.

\bibitem[Nguyen et~al.(2017)Nguyen, Clune, Bengio, Dosovitskiy, and Yosinski]{nguyen2017plug}
Anh Nguyen, Jeff Clune, Yoshua Bengio, Alexey Dosovitskiy, and Jason Yosinski.
\newblock Plug \& play generative networks: Conditional iterative generation of images in latent space.
\newblock In \emph{Proceedings of the IEEE conference on computer vision and pattern recognition}, pp.\  4467--4477, 2017.

\bibitem[Nguyen et~al.(2021)Nguyen, Kim, and Nguyen]{nguyen2021effectiveness}
Giang Nguyen, Daeyoung Kim, and Anh Nguyen.
\newblock The effectiveness of feature attribution methods and its correlation with automatic evaluation scores.
\newblock \emph{Advances in Neural Information Processing Systems (NeurIPS)}, 2021.

\bibitem[Nguyen et~al.(2022)Nguyen, Taesiri, and Nguyen]{nguyen2022visual}
Giang Nguyen, Mohammad~Reza Taesiri, and Anh Nguyen.
\newblock Visual correspondence-based explanations improve ai robustness and human-ai team accuracy.
\newblock \emph{arXiv preprint arXiv:2208.00780}, 1, 2022.

\bibitem[Novello et~al.(2022)Novello, Fel, and Vigouroux]{novello2022making}
Paul Novello, Thomas Fel, and David Vigouroux.
\newblock Making sense of dependence: Efficient black-box explanations using dependence measure.
\newblock In \emph{Advances in Neural Information Processing Systems (NeurIPS)}, 2022.

\bibitem[Olah et~al.(2017)Olah, Mordvintsev, and Schubert]{olah2017feature}
Chris Olah, Alexander Mordvintsev, and Ludwig Schubert.
\newblock Feature visualization.
\newblock \emph{Distill}, 2017.

\bibitem[Paszke et~al.(2019)Paszke, Gross, Massa, Lerer, Bradbury, Chanan, Killeen, Lin, Gimelshein, Antiga, et~al.]{paszke2019pytorch}
Adam Paszke, Sam Gross, Francisco Massa, Adam Lerer, James Bradbury, Gregory Chanan, Trevor Killeen, Zeming Lin, Natalia Gimelshein, Luca Antiga, et~al.
\newblock Pytorch: An imperative style, high-performance deep learning library.
\newblock \emph{Advances in neural information processing systems}, 32, 2019.

\bibitem[Petsiuk et~al.(2018)Petsiuk, Das, and Saenko]{petsiuk2018rise}
Vitali Petsiuk, Abir Das, and Kate Saenko.
\newblock Rise: Randomized input sampling for explanation of black-box models.
\newblock In \emph{Proceedings of the British Machine Vision Conference (BMVC)}, 2018.

\bibitem[Poyiadzi et~al.(2020)Poyiadzi, Sokol, Santos-Rodriguez, De~Bie, and Flach]{poyiadzi2020face}
Rafael Poyiadzi, Kacper Sokol, Raul Santos-Rodriguez, Tijl De~Bie, and Peter Flach.
\newblock Face: feasible and actionable counterfactual explanations.
\newblock In \emph{Proceedings of the AAAI/ACM Conference on AI, Ethics, and Society}, pp.\  344--350, 2020.

\bibitem[Rebuffi et~al.(2020)Rebuffi, Fong, Ji, and Vedaldi]{rebuffi2020there}
Sylvestre-Alvise Rebuffi, Ruth Fong, Xu~Ji, and Andrea Vedaldi.
\newblock There and back again: Revisiting backpropagation saliency methods.
\newblock In \emph{Proceedings of the IEEE Conference on Computer Vision and Pattern Recognition (CVPR)}, 2020.

\bibitem[Ribeiro et~al.(2016)Ribeiro, Singh, and Guestrin]{ribeiro2016i}
Marco~Tulio Ribeiro, Sameer Singh, and Carlos Guestrin.
\newblock "why should i trust you?": Explaining the predictions of any classifier.
\newblock In \emph{Knowledge Discovery and Data Mining (KDD)}, 2016.

\bibitem[Santurkar et~al.(2019)Santurkar, Ilyas, Tsipras, Engstrom, Tran, and Madry]{santurkar2019image}
Shibani Santurkar, Andrew Ilyas, Dimitris Tsipras, Logan Engstrom, Brandon Tran, and Aleksander Madry.
\newblock Image synthesis with a single (robust) classifier.
\newblock \emph{Advances in Neural Information Processing Systems}, 32, 2019.

\bibitem[Selvaraju et~al.(2017{\natexlab{a}})Selvaraju, Cogswell, Das, Vedantam, Parikh, and Batra]{GradCAM}
Ramprasaath~R Selvaraju, Michael Cogswell, Abhishek Das, Ramakrishna Vedantam, Devi Parikh, and Dhruv Batra.
\newblock Grad-cam: Visual explanations from deep networks via gradient-based localization.
\newblock In \emph{Proceedings of the IEEE international conference on computer vision}, pp.\  618--626, 2017{\natexlab{a}}.

\bibitem[Selvaraju et~al.(2017{\natexlab{b}})Selvaraju, Cogswell, Das, Vedantam, Parikh, and Batra]{Selvaraju_2019}
Ramprasaath~R. Selvaraju, Michael Cogswell, Abhishek Das, Ramakrishna Vedantam, Devi Parikh, and Dhruv Batra.
\newblock Grad-cam: Visual explanations from deep networks via gradient-based localization.
\newblock In \emph{Proceedings of the IEEE International Conference on Computer Vision (ICCV)}, 2017{\natexlab{b}}.

\bibitem[Selvaraju et~al.(2017{\natexlab{c}})Selvaraju, Cogswell, Das, Vedantam, Parikh, and Batra]{selvaraju2017grad}
Ramprasaath~R Selvaraju, Michael Cogswell, Abhishek Das, Ramakrishna Vedantam, Devi Parikh, and Dhruv Batra.
\newblock Grad-cam: Visual explanations from deep networks via gradient-based localization.
\newblock In \emph{Proceedings of the IEEE international conference on computer vision}, pp.\  618--626, 2017{\natexlab{c}}.

\bibitem[Shen \& Huang(2020)Shen and Huang]{shen2020useful}
Hua Shen and Ting-Hao Huang.
\newblock How useful are the machine-generated interpretations to general users? a human evaluation on guessing the incorrectly predicted labels.
\newblock In \emph{Proceedings of the AAAI Conference on Human Computation and Crowdsourcing}, volume~8, pp.\  168--172, 2020.

\bibitem[Shrikumar et~al.(2017)Shrikumar, Greenside, and Kundaje]{shrikumar2017learning}
Avanti Shrikumar, Peyton Greenside, and Anshul Kundaje.
\newblock Learning important features through propagating activation differences.
\newblock In \emph{Proceedings of the International Conference on Machine Learning (ICML)}, 2017.

\bibitem[Simonyan et~al.(2013)Simonyan, Vedaldi, and Zisserman]{simonyan2013deep}
Karen Simonyan, Andrea Vedaldi, and Andrew Zisserman.
\newblock Deep inside convolutional networks: Visualising image classification models and saliency maps.
\newblock In \emph{Workshop, Proceedings of the International Conference on Learning Representations (ICLR)}, 2013.

\bibitem[Sixt et~al.(2022)Sixt, Schuessler, Popescu, Wei{\ss}, and Landgraf]{sixt2022users}
Leon Sixt, Martin Schuessler, Oana-Iuliana Popescu, Philipp Wei{\ss}, and Tim Landgraf.
\newblock Do users benefit from interpretable vision? a user study, baseline, and dataset.
\newblock \emph{arXiv preprint arXiv:2204.11642}, 2022.

\bibitem[Slack et~al.(2020)Slack, Hilgard, Jia, Singh, and Lakkaraju]{slack2020fooling}
Dylan Slack, Sophie Hilgard, Emily Jia, Sameer Singh, and Himabindu Lakkaraju.
\newblock Fooling lime and shap: Adversarial attacks on post hoc explanation methods.
\newblock In \emph{Proceedings of the AAAI/ACM Conference on AI, Ethics, and Society}, pp.\  180--186, 2020.

\bibitem[Smilkov et~al.(2017)Smilkov, Thorat, Kim, Viégas, and Wattenberg]{smilkov2017smoothgrad}
Daniel Smilkov, Nikhil Thorat, Been Kim, Fernanda Viégas, and Martin Wattenberg.
\newblock Smoothgrad: removing noise by adding noise.
\newblock In \emph{Workshop on Visualization for Deep Learning, Proceedings of the International Conference on Machine Learning (ICML)}, 2017.

\bibitem[Springenberg et~al.(2014{\natexlab{a}})Springenberg, Dosovitskiy, Brox, and Riedmiller]{guided-backprop}
Jost~Tobias Springenberg, Alexey Dosovitskiy, Thomas Brox, and Martin Riedmiller.
\newblock Striving for simplicity: The all convolutional net.
\newblock \emph{arXiv preprint arXiv:1412.6806}, 2014{\natexlab{a}}.

\bibitem[Springenberg et~al.(2014{\natexlab{b}})Springenberg, Dosovitskiy, Brox, and Riedmiller]{springenberg2014striving}
Jost~Tobias Springenberg, Alexey Dosovitskiy, Thomas Brox, and Martin Riedmiller.
\newblock Striving for simplicity: The all convolutional net.
\newblock In \emph{Workshop Proceedings of the International Conference on Learning Representations (ICLR)}, 2014{\natexlab{b}}.

\bibitem[Srinivas \& Fleuret(2019)Srinivas and Fleuret]{srinivas2019full}
Suraj Srinivas and Fran{\c{c}}ois Fleuret.
\newblock Full-gradient representation for neural network visualization.
\newblock \emph{Advances in neural information processing systems}, 32, 2019.

\bibitem[Sundararajan et~al.(2017)Sundararajan, Taly, and Yan]{sundararajan2017axiomatic}
Mukund Sundararajan, Ankur Taly, and Qiqi Yan.
\newblock Axiomatic attribution for deep networks.
\newblock In \emph{Proceedings of the International Conference on Machine Learning (ICML)}, 2017.

\bibitem[Szegedy et~al.(2013)Szegedy, Zaremba, Sutskever, Bruna, Erhan, Goodfellow, and Fergus]{szegedy2013intriguing}
Christian Szegedy, Wojciech Zaremba, Ilya Sutskever, Joan Bruna, Dumitru Erhan, Ian Goodfellow, and Rob Fergus.
\newblock Intriguing properties of neural networks.
\newblock \emph{arXiv preprint arXiv:1312.6199}, 2013.

\bibitem[Szegedy et~al.(2015)Szegedy, Liu, Jia, Sermanet, Reed, Anguelov, Erhan, Vanhoucke, and Rabinovich]{vgg}
Christian Szegedy, Wei Liu, Yangqing Jia, Pierre Sermanet, Scott Reed, Dragomir Anguelov, Dumitru Erhan, Vincent Vanhoucke, and Andrew Rabinovich.
\newblock Going deeper with convolutions.
\newblock In \emph{Proceedings of the IEEE conference on computer vision and pattern recognition}, pp.\  1--9, 2015.

\bibitem[Tsipras et~al.(2018)Tsipras, Santurkar, Engstrom, Turner, and Madry]{tsipras2018robustness}
Dimitris Tsipras, Shibani Santurkar, Logan Engstrom, Alexander Turner, and Aleksander Madry.
\newblock Robustness may be at odds with accuracy.
\newblock \emph{arXiv preprint arXiv:1805.12152}, 2018.

\bibitem[Tyka(2016)]{tyka2016class}
Mike Tyka.
\newblock Class visualization with bilateral filters. 2016.
\newblock \emph{URL: https://mtyka. github. io/deepdream/2016/02/05/bilateral-class-vis. html}, 2\penalty0 (3), 2016.

\bibitem[Verma et~al.(2020)Verma, Boonsanong, Hoang, Hines, Dickerson, and Shah]{verma2020counterfactual}
Sahil Verma, Varich Boonsanong, Minh Hoang, Keegan~E Hines, John~P Dickerson, and Chirag Shah.
\newblock Counterfactual explanations and algorithmic recourses for machine learning: A review.
\newblock \emph{arXiv preprint arXiv:2010.10596}, 2020.

\bibitem[Wang et~al.(2020)Wang, Wang, Du, Yang, Zhang, Ding, Mardziel, and Hu]{wang2020score}
Haofan Wang, Zifan Wang, Mengnan Du, Fan Yang, Zijian Zhang, Sirui Ding, Piotr Mardziel, and Xia Hu.
\newblock Score-cam: Score-weighted visual explanations for convolutional neural networks.
\newblock In \emph{Proceedings of the IEEE/CVF conference on computer vision and pattern recognition workshops}, pp.\  24--25, 2020.

\bibitem[Wei et~al.(2015)Wei, Zhou, Torrabla, and Freeman]{wei2015understanding}
Donglai Wei, Bolei Zhou, Antonio Torrabla, and William Freeman.
\newblock Understanding intra-class knowledge inside cnn.
\newblock \emph{arXiv preprint arXiv:1507.02379}, 2015.

\bibitem[Yang et~al.(2020)Yang, Qiu, Song, Tao, and Wang]{yang2020learning}
Yiding Yang, Jiayan Qiu, Mingli Song, Dacheng Tao, and Xinchao Wang.
\newblock Learning propagation rules for attribution map generation.
\newblock In \emph{Computer Vision--ECCV 2020: 16th European Conference, Glasgow, UK, August 23--28, 2020, Proceedings, Part XX 16}, pp.\  672--688. Springer, 2020.

\bibitem[Zeiler \& Fergus(2014{\natexlab{a}})Zeiler and Fergus]{zeiler2013visualizing}
Matthew~D Zeiler and Rob Fergus.
\newblock Visualizing and understanding convolutional networks.
\newblock In \emph{Proceedings of the IEEE European Conference on Computer Vision (ECCV)}, 2014{\natexlab{a}}.

\bibitem[Zeiler \& Fergus(2014{\natexlab{b}})Zeiler and Fergus]{zeiler2014visualizing}
Matthew~D Zeiler and Rob Fergus.
\newblock Visualizing and understanding convolutional networks.
\newblock In \emph{Proceedings of the IEEE European Conference on Computer Vision (ECCV)}, 2014{\natexlab{b}}.

\bibitem[Zhou et~al.(2016)Zhou, Khosla, Lapedriza, Oliva, and Torralba]{zhou2016learning}
Bolei Zhou, Aditya Khosla, Agata Lapedriza, Aude Oliva, and Antonio Torralba.
\newblock Learning deep features for discriminative localization.
\newblock In \emph{Proceedings of the IEEE conference on computer vision and pattern recognition}, pp.\  2921--2929, 2016.

\bibitem[Zhou et~al.(2018)Zhou, Sun, Bau, and Torralba]{Zhou_2018_ECCV}
Bolei Zhou, Yiyou Sun, David Bau, and Antonio Torralba.
\newblock Interpretable basis decomposition for visual explanation.
\newblock In \emph{Proceedings of the European Conference on Computer Vision (ECCV)}, September 2018.

\bibitem[Zimmermann et~al.(2021)Zimmermann, Borowski, Geirhos, Bethge, Wallis, and Brendel]{zimmermann2021well}
Roland~S Zimmermann, Judy Borowski, Robert Geirhos, Matthias Bethge, Thomas Wallis, and Wieland Brendel.
\newblock How well do feature visualizations support causal understanding of cnn activations?
\newblock \emph{Advances in Neural Information Processing Systems}, 34:\penalty0 11730--11744, 2021.

\bibitem[Zintgraf et~al.(2017)Zintgraf, Cohen, Adel, and Welling]{zintgraf2017visualizing}
Luisa~M Zintgraf, Taco~S Cohen, Tameem Adel, and Max Welling.
\newblock Visualizing deep neural network decisions: Prediction difference analysis.
\newblock In \emph{Proceedings of the International Conference on Learning Representations (ICLR)}, 2017.

\end{thebibliography}
